\documentclass[12pt]{article}

\usepackage{float}

\usepackage{multirow}

\usepackage{tikz}

\usetikzlibrary{calc,backgrounds,arrows,matrix}

\usepackage{enumerate}
\usepackage{listings}

\usepackage{color}
\definecolor{lightblue}{rgb}{0,0.2,0.5}

\usepackage[colorlinks=true, urlcolor=blue,linkcolor=blue, citecolor=lightblue]{hyperref}

\usepackage[round,sort,comma,numbers]{natbib}

\bibpunct{\textcolor{lightblue}{(}}{\textcolor{lightblue}{)}}{,}{a}{}{;}

\usepackage{amssymb,amsmath}

\usepackage{graphicx}
\usepackage{subfigure}

\usepackage{xcolor}
\usepackage{mathrsfs}

\usepackage{mathtools}
\usepackage{accents}

\DeclareMathAlphabet{\eufrak}{U}{}{}{}

\SetMathAlphabet\eufrak{normal}{U}{euf}{m}{n}
\SetMathAlphabet\eufrak{bold}{U}{euf}{b}{n}

\allowdisplaybreaks

 \def\qu{{\mathord{\mathbb Z}}}

 \def\sZZ{{\rm Z\kern-.45em{}Z}}

 \def\sQQ{{\kern 0.27em \vrule height1.45ex width0.03em depth0em
           \kern-0.30em \rm Q}}
 \def\qu{{\mathchoice
         {\sQQ}
         {\sQQ}
   {\kern 0.225em \vrule height1.05ex width0.025em depth0em \kern-0.25em \rm Q}
   {\kern 0.180em \vrule height0.78ex width0.020em depth0em \kern-0.20em \rm Q}
         }}
 \def\sGG{{\kern 0.27em \vrule height1.45ex width0.03em depth0em
           \kern-0.30em \rm G}}
 \def\gg{{\mathchoice
         {\sGG}
         {\sGG}
   {\kern 0.225em \vrule height1.05ex width0.025em depth0em \kern-0.25em \rm G}
   {\kern 0.180em \vrule height0.78ex width0.020em depth0em \kern-0.20em \rm G}
         }}

 \newtheorem{prop}{Proposition}[section]
 
 \newtheorem{definition}[prop]{Definition}
 
 \newtheorem{theorem}[prop]{Theorem}
 \newtheorem{algo}[prop]{Algorithm}
 \newtheorem{remark}[prop]{Remark}
 
 \newtheorem{assumption}{Assumption}

\numberwithin{equation}{section}

\usepackage{textcomp}

\usepackage{accsupp}    

\newcommand{\noncopynumber}[1]{
    \BeginAccSupp{method=escape,ActualText={}}
    #1
    \EndAccSupp{}
}

\newcommand{\abs}[1]{\lvert#1\rvert}
\newcommand{\norm}[1]{\lVert#1\rVert}

 \def\P{{\mathord{\mathbb P}}}

\newcommand{\re}{\mathrm{e}}

\newcommand{\ubar}[1]{\smash{\underline {#1}}}

 \newcounter{hyp}
\usepackage{aeguill}

\newenvironment{Proof}{\removelastskip\par\medskip \noindent{\em Proof.} \rm}{\penalty-20\null\hfill$\square$\par\medbreak}

\def\bprf{\begin{Proof}}
\def\nprf{\end{Proof}}
\def\bdes{\begin{description}}
\def\ndes{\end{description}}

\newtheorem{thm}{Theorem}[section]

\def\bdef{\begin{defn}}
\def\ndef{\end{defn}}
\def\bthm{\begin{thm}}
\def\nthm{\end{thm}}
\def\bprop{\begin{prop}}
\def\nprop{\end{prop}}
\def\brmk{\begin{remark}}
\def\nrmk{\end{remark}}
\def\bexa{\begin{exa}}
\def\nexa{\end{exa}}
\def\blem{\begin{lem}}
\def\nlem{\end{lem}}
\def\bcor{\begin{cor}}
\def\ncor{\end{cor}}
\def\bexe{\begin{exe}}
\def\nexe{\end{exe}}

\def\esssup{{\mathrm{ess~{\rm sup}}}}

\newcommand{\E}{\mathbb{E}}

\newcommand{\real}{\mathbb{R}}

\def\diag{\mathop{\hbox{\rm Diag}}\nolimits}

\def\tr{\mathop{\hbox{\rm tr}}\nolimits}

\def\og{\leavevmode\raise.3ex
     \hbox{$\scriptscriptstyle\langle\!\langle$~}}
\def\fg{\leavevmode\raise.3ex
     \hbox{~$\!\scriptscriptstyle\,\rangle\!\rangle$}~}

\title{\Huge
 A deep learning approach to the probabilistic numerical solution of
 path-dependent partial differential equations
}

\author{
 Jiang Yu Nguwi\footnote{\href{mailto:nguw0003@e.ntu.edu.sg}{nguw0003@e.ntu.edu.sg}
 }
 \qquad Nicolas Privault\footnote{
\href{mailto:nprivault@ntu.edu.sg}{nprivault@ntu.edu.sg}
 }
 \\
  \small
Division of Mathematical Sciences
\\
\small
School of Physical and Mathematical Sciences
\\
\small
Nanyang Technological University
\\
\small
21 Nanyang Link, Singapore 637371
}

\allowdisplaybreaks

\usepackage{empheq}
\usepackage{bm}

\usepackage{bbm}

\makeatletter
\newcommand*\rel@kern[1]{\kern#1\dimexpr\macc@kerna}
\newcommand*\widebar[1]{
  \begingroup
  \def\mathaccent##1##2{
    \rel@kern{0.8}
    \overline{\rel@kern{-0.8}\macc@nucleus\rel@kern{0.2}}
    \rel@kern{-0.2}
  }
  \macc@depth\@ne
  \let\math@bgroup\@empty \let\math@egroup\macc@set@skewchar
  \mathsurround\z@ \frozen@everymath{\mathgroup\macc@group\relax}
  \macc@set@skewchar\relax
  \let\mathaccentV\macc@nested@a
  \macc@nested@a\relax111{#1}
  \endgroup
}
\makeatother

\let\oldcitet=\citet
\let\oldcitep=\citep
\renewcommand{\cite}[1]{\textcolor[rgb]{0,0,1}{\oldcitet{#1}}}
\renewcommand{\citet}[1]{\textcolor[rgb]{0,0,1}{\oldcitet{#1}}}
\renewcommand{\citep}[1]{\textcolor[rgb]{0,0,1}{\oldcitep{#1}}}

\begin{document}
\maketitle

\baselineskip0.6cm

\vspace{-0.6cm}

\begin{abstract}
\medskip
Recent work on Path-Dependent Partial Differential Equations (PPDEs) has shown that PPDE solutions can be approximated by a probabilistic representation, implemented in the literature by the estimation of conditional expectations using regression. However, a limitation of this approach is to require the selection of a basis in a function space. In this paper, we overcome this limitation by the use of deep learning methods, and we show that this setting allows for the derivation of error bounds on the approximation of conditional expectations. Numerical examples based on a two-person zero-sum game, as well as on Asian and barrier option pricing, are presented. In comparison with other deep learning approaches, our algorithm appears to be more accurate, especially in large dimensions.
\end{abstract}

    

\baselineskip0.7cm

\parskip-0.1cm

\section{Introduction}
 Fully nonlinear PPDEs of the form
\begin{equation}
    \label{eq:ppde}
    \begin{cases}
      \displaystyle
      \partial_t u (t, \omega) + b(t, \omega) \cdot \partial_\omega u(t, \omega)
	+ \frac{1}{2}\sigma\sigma^\top(t, \omega):\partial_{\omega\omega} u(t, \omega)
  + F\left( \cdot , u, \sigma^\top \partial_\omega u, \sigma^\top
	\partial^2_{\omega\omega}u\sigma\right)(t, \omega) = 0,
    \medskip
    \\
     u(T, \omega) = g(\omega),
    \end{cases}
\end{equation}
 have been introduced in
 \cite{peng2011note}
 and their well-posedness in the sense of viscosity solutions
 have been studied in
 \cite{ekren2014viscosity,ekren2016viscosity1,ekren2016viscosity2}.
 Here, $\omega$ is in the set
 of $\real^d$-valued continuous paths,
 $b (t,\omega )$ is $\real^d$-valued,
 $\sigma (t,\omega )$ takes values
 in the space of invertible $d\times d$ matrices,
 and $F$ is a real-valued function
 on $\real_+\times \real^3$ satisfying Assumption~\ref{basic assumptions} below.
 The precise meaning of the partial derivatives
$\partial_t u (t, \omega)$, $\partial_\omega u(t, \omega)$
and $\partial_{\omega\omega} u(t, \omega)$,
which are connected to
the horizontal and vertical derivatives
 of functional It\^o calculus introduced in \cite{dupire2009functional},
 will be discussed in Section~\ref{sec:ppde}.

 \medskip

 PPDEs of the type \eqref{eq:ppde} have recently been
 the object of increased attention
 due to their ability to model control and pricing problems
 in a non-Markovian setting, see e.g.
\cite{tang2015},
\cite{jacquier2019deep},
\cite{viens2019martingale}.

\medskip

Nevertheless, a large class of PPDE is not analytically solvable,
and one has to rely on the numerical solution.
In \cite{ren2017convergence}, a probabilistic scheme based on
\cite{fahim2011probabilistic} has been proposed, and was proved
to converge to the viscosity solution of PPDE.
However, its practical implementation
is far from trivial due to the presence of the conditional expectation.
The suggestion of \cite{ren2017convergence}
to use regression as in \cite{gobet2005regression}
relies on a careful basis function choice, which may not always be possible,
see the discussion at the end of Section~\ref{sec:ppde}.

\medskip

Neural networks methods for PDEs have been introduced independently in
\cite{han2018solving} and \cite{sirignano2018dgm}
using backward stochastic differential equations and
the Galerkin method respectively, see also
\cite{beck2019deep,hure2019some}
for other variants of deep learning-based numerical solutions.

\medskip

 A deep neural network algorithm for the numerical solution of PPDEs
 has also been proposed in \cite{saporito2020pdgm}
 by applying Long Short-Term Memory (LSTM) networks
 in the framework of the deep Galerkin method for PDEs,
 see \cite{sirignano2018dgm}.
On the other hand, \cite{sabate2020solving} propose to combine
the LSTM network and the path signature to solve the linear PPDE.
Unlike regression methods, deep learning algorithms do
not rely on the choice of a basis.

\medskip

In this paper, we propose a deep learning approach
 to the implementation of the probabilistic scheme of \cite{ren2017convergence}
 for the numerical solution of
 fully nonlinear PPDEs of the form \eqref{eq:ppde}.
 The main idea of Algorithm~\ref{algo:deep_ppde}
is based on the $L^2$ minimality property
of the conditional expectations,
which allows us to transform the conditional expectation
in the probabilitstic scheme into an optimization problem
\eqref{eq:probabilistic_scheme} that can be solved using neural networks.

\medskip

Additionally, we propose an error bound of the Algorithm~\ref{algo:deep_ppde}
in Proposition~\ref{theo:deep_scheme_nonlinear},
which is used to prove its convergence 
in Theorem~\ref{corollary_convergence}.
In Section~\ref{sec:numerical},
we detail the implementation of our algorithm,
followed by numerical examples of
two-person zero-sum game, Asian and barrier options pricing.
In the numerical comparisons,
 Algorithm~\ref{algo:deep_ppde} appears more accurate
than the deep learning algorithms
of \cite{saporito2020pdgm} and \cite{sabate2020solving}.

\medskip

This paper is organized as follows.
The necessary preliminaries on PPDEs are stated in Section~\ref{sec:ppde},
and the probabilistic schemes
introduced in \cite{fahim2011probabilistic}
for PDEs and in \cite{ren2017convergence}
for PPDEs are reviewed in Section~\ref{s3}.
In Section~\ref{sec:deep}
we present our main deep learning algorithm, and
derive its error bound and convergence respectively in
Proposition~\ref{theo:deep_scheme_nonlinear}
and Theorem~\ref{corollary_convergence}.
Numerical implementation and examples are presented
in Section~\ref{sec:numerical}.

\section{Viscosity solutions of Path-dependent PDEs} 
\label{sec:ppde}
 Fix $T > 0$, $x_0 \in \real^d$,
 let $\Omega = {\cal C}_{x_0} ([0,T]; \real^d)$
 denote the set
 of $\real^d$-valued continuous paths $\omega$ started at $\omega_0 = x_0$
and let $\Theta := [0,T] \times \Omega$.
We let $B=(B_t)_{t\in [0,T]}$
denote the $\real^d$-valued canonical process on $\Omega$,
while $\mathbb{F} = ( \mathcal{F}_t )_{0 \leq t \leq T}$
denotes the canonical filtration
generated by $(B_t)_{t\in [0,T]}$, and
$\P_0$ is the Wiener measure on $( \Omega , \mathbb{F} )$.
A natural metric $d$ on $\Theta$ is defined as
\begin{equation*}
	d\left( (t,\omega), (t', \omega') \right) =
	\abs{t-t'} + \norm{ \omega_{t\land\cdot} - \omega'_{t'\land\cdot} },
 \qquad (t,\omega),(t',\omega') \in \Theta,
\end{equation*}
where
$\displaystyle \norm{\omega} := \sup\limits_{t \in [0,T]} \norm{\omega_t}_d$,
$\omega \in \Omega$, and
  $\norm{\ \! \cdot \ \! }_d$ denotes the Euclidean norm on
$\real^d$.
Next, we define $\mathcal{P}$ as the set of
probability measures $\mathbb{P}$ such that
the canonical process
$$ B_t = A^\mathbb{P}_t + M^\mathbb{P}_t
= \int_0^t \mu^\mathbb{P}_s ds + M^\mathbb{P}_t, \qquad t\in [0,T],
$$
is a semimartingale,
where $(A^\mathbb{P})_{t\in [0,T]}$ is a finite variation process
and $(M^\mathbb{P})_{t\in [0,T]}$ is a continuous martingale,
so that the quadratic variation
$$
\langle M^\mathbb{P} \rangle_t = \int_0^t a^\mathbb{P}_s ds, \qquad t\in [0,T],
$$
 taking values in the space $\mathbb{S}^d$ of $d \times d$ symmetric matrices
is absolutely continuous with respect to
the Lebesgue measure on $[0,T]$, and
\begin{equation*}
  \sup\limits_{t \in[0,T]} \norm{\mu^\mathbb{P}_t}_d\leq L,
  \qquad
	\frac{1}{2}\sup\limits_{t \in[0,T]} {\rm Tr} \big(a^\mathbb{P}_t\big) \leq L,
	\quad \mathbb{P}\text{-a.s.}, 
\end{equation*}
 where $L > 0$ is fixed throughout the paper.
 We denote by $v_1 \cdot v_2$ the dot product of
$v_1, v_2 \in \real^d$, let $\bm{I}_d$
be the identity matrix in $\mathbb{S}^d$,
 let $A:B = {\rm Tr} (AB)$,
 and let $\bm{1}_d=(1,\ldots , 1)$ represent the all-ones vector in $\real^d$.
The following definition makes sense of
the partial derivatives
$\partial_\omega u(t, \omega)$,
$\partial_{\omega\omega} u(t, \omega)$
        used in the PPDE \eqref{eq:ppde}
 as {\em path derivatives} on $\Theta$,
 see \cite{ekren2016viscosity1}.
\begin{definition}
\nonumber 
	We say that
    $u \in C^{1,2}(\Theta)$
	if $(t, \omega ) \mapsto u_t(\omega ) \in \real$ is continuous
        on $(\Theta , d)$
	and there exists continuous processses
        $(t, \omega ) \mapsto \partial_t u \in \real$,
	$(t, \omega ) \mapsto \partial_\omega u_t \in \real^d$,
        $(t, \omega ) \mapsto \partial_{\omega\omega} u_t \in \mathbb{S}^d$,
        continuous         on $(\Theta , d)$, and such that
	\begin{equation}
          \nonumber
	  du_t = \partial_t u dt + \frac{1}{2} \partial_{\omega\omega} u_t:d\langle B \rangle_t
		+ \partial_\omega u_t \cdot dB_t, \quad \P\text{-a.s., for all } \P \in \mathcal{P}.
	\end{equation}
\end{definition}

\noindent
As in e.g.
\cite{ren2017convergence} and
\cite{fahim2011probabilistic},
we assume that the coefficients of
\eqref{eq:ppde}
 satisfy the following conditions
throughout the paper.
\begin{assumption}
\label{basic assumptions}
\begin{enumerate}[i)]
\item $\sigma (t, \omega) \in \mathbb{S}^d$ is invertible
  for all $(t, \omega)\in \Theta$, and
		$b(t, \omega)$, $\sigma (t, \omega)$ satisfy
		\begin{equation*}
			\sup\limits_{(t, \omega) \neq (t', \omega')}
			\frac{\norm{b(t, \omega) - b(t', \omega')}_d}
			{\abs{t-t'}^{1/2} + \norm{\omega_{t \land \cdot}
				- \omega'_{t \land \cdot}}}
			+ \sup\limits_{(t, \omega) \neq (t', \omega')}
            \frac{\norm{\sigma(t, \omega) - \sigma(t', \omega')}_{d\times d}}
			{\abs{t-t'}^{1/2} + \norm{\omega_{t \land \cdot}
				- \omega'_{t \land \cdot}}}
			< \infty,
		\end{equation*}
  where
  $\norm{ \cdot }_{d\times d}$ denotes the Frobenius norm on $\mathbb{S}^d$.
	      \item
                \label{part2}
                $\omega \mapsto g(\omega)$ is bounded Lipschitz on $\Omega$,
	\item $F(t, \omega, u, z, \gamma)$ is
	  continuous in $(t, \omega, u, z, \gamma)\in \Theta \times \real \times \real^d \times \mathbb{S}^d$ and
	  non-decreasing in $\gamma \in \mathbb{S}^d$
          for the positive semidefinite order $\leq_{\rm psd}$.
	\item
          \label{aaa}
          $F (t, \omega, u, z, \gamma)$ is
	  Lipschitz with respect to $(\omega, u, z, \gamma)
          \in \Omega \times \real \times \real^d \times \mathbb{S}^d$
		uniformly in $t \in [0,T]$, and
		$\sup\limits_{(t, \omega) \in \Theta}
		\abs{F(t, \omega, 0, 0, 0)} < \infty$.
	      \item $F(t, \omega, u, z, \gamma)$ is elliptic,
                i.e., $F(t, \omega, u, z, \gamma) \leq F(t, \omega, u, z, \gamma ')$
                for $\gamma \leq \gamma'$,
                and
		satisfies
                $\displaystyle \frac{\partial F}{\partial \gamma} (t, \omega, u, z, \gamma)\leq_{\rm psd} \bm{I}_d$
				for a.e. $(t, \omega, u, z, \gamma)\in \Theta \times \real \times \real^d \times \mathbb{S}^d$.
	      \item $\displaystyle
                \frac{\partial F}{\partial z} (t, \omega, u, z, \gamma) \in
		     {\rm Image} \left(\sigma
                     \frac{\partial F}{\partial \gamma} \sigma^\top (t, \omega, u, z, \gamma)\right)$
                for all
		$(t, \omega, u, z, \gamma) \in
                \Theta \times \real \times \real^d \times \mathbb{S}^d$, and
		$$
                \esssup_{(t, \omega, u, z, \gamma)}
		\left| \left(\frac{\partial F}{\partial z} \right)^\top
		\left(\sigma \frac{\partial F}{\partial \gamma} \sigma^\top\right)^{-1}
		  \frac{\partial F}{\partial z} (t, \omega, u, z, \gamma)\right| < \infty.
                $$
\end{enumerate}
\end{assumption}
\noindent
In general, $u$ may not be smooth enough to ensure
the existence of the classical solution for \eqref{eq:ppde},
hence we rely on the weaker notion of viscosity solution.
For this, we will use the shift operations
$$
	(\omega \otimes_t \omega')_s := \omega_s \mathbbm{1}_{[0,t]}(s) +
			(\omega'_{s-t} - x_0 + \omega_t) \mathbbm{1}_{(t,T]}(s),
  \quad
  \omega, \omega' \in \Omega,
$$
        and, for $u:\Theta \to \real$,
$$
        u^{t,\omega}(s, \omega'):= u(t+s, \omega \otimes_t \omega'),
        \quad
        \omega, \omega' \in \Omega.
$$
 Next, for $u$ in the set BUC$(\Theta)$
of all bounded and uniformly continuous functions $u:\Theta \rightarrow \real$
 on $(\Theta , d)$
 we define the sets of test functions
$$
\overline{\mathcal{A}}u(t, \omega) := \Big\{
		\varphi \in C^{1,2}(\Theta) :
		(\varphi - u^{t,\omega})_0 = 0 =\sup\limits_{\tau \in \mathcal{T}_{H_\delta}}
		\overline{\mathcal{E}} \left[ (\varphi - u^{t, \omega})_\tau \right]
	        \Big\}
                $$
                and
                $$
	\ubar{\mathcal{A}}u(t, \omega) := \Big\{
		\varphi \in C^{1,2}(\Theta) :
		(\varphi - u^{t,\omega})_0 = 0 = \inf\limits_{\tau \in \mathcal{T}_{H_\delta}}
		\ubar{\mathcal{E}} \left[ (\varphi - u^{t, \omega})_\tau \right]
	\Big\},
$$
        $(t,\omega) \in \Theta$, where
$H_\delta(\omega') := \delta \land \inf\{ s \ge 0: \abs{\omega'_s} \ge \delta \}$,
$\mathcal{T}_{H_\delta}$ is the set of all
$\mathbb{F}$-stopping times taking values in $[0,H_\delta]$,
$\overline{\mathcal{E}} [ \ \! \cdot \ \! ] := \sup\limits_{\P \in \mathcal{P} } \E^\P[ \ \! \cdot \ \! ]$,
and $\ubar{\mathcal{E}} [ \ \! \cdot \ \! ] := \inf\limits_{\P \in \mathcal{P} } \E^\P[ \ \! \cdot \ \! ]$.
\\
The following definition makes sense of
the viscosity solution of the PPDE \eqref{eq:ppde}.
\begin{definition}
	 \begin{enumerate}[i)]
		 \item $u \in $ BUC$(\Theta)$ is a viscosity subsolution
			 (resp. supersolution) of the PPDE \eqref{eq:ppde}
			 if for any $(t,\omega) \in \Theta$ and
			 for all
			 $\varphi \in \ubar{\mathcal{A}} u(t, \omega)$,
			 (resp. $\varphi \in \bar{\mathcal{A}} u(t, \omega)$),
			 we have
                         $$
                         \! \! \! \! \!
                        \partial_t \varphi (t, \omega)
			+ b(t, \omega)
			\cdot \partial_\omega \varphi(t, \omega)
			+ \frac{1}{2}\sigma\sigma^\top(t, \omega)
			:\partial_{\omega\omega} \varphi(t, \omega)
						+ F\left(\ \! \cdot \ \! , \varphi,
				\sigma^\top \partial_\omega \varphi,
				\sigma^\top
				\partial^2_{\omega\omega}\varphi
				\sigma\right)(t, \omega)
			\ge 0,
$$
                      resp. $\leq 0$.
	              \item $u$ is a viscosity solution of
		 the PPDE \eqref{eq:ppde} if
		 it is both a viscosity subsolution and
		 a viscosity supersolution of \eqref{eq:ppde}.
	\end{enumerate}
\end{definition}
\section{Probabilistic numerical solution}
\label{s3}
Next, we consider the probabilistic scheme
introduced by \cite{fahim2011probabilistic}
for PDEs,
and later generalized to PPDEs
by \cite{ren2017convergence}.
For a given $N \ge 1$ and $h = T/N$, define the random variable
\begin{equation}
	\label{eq:sde_discrete}
	X^{(t,\omega)}_h :=
\begin{pmatrix}
    x_0     \\
    x_0 + b(t, \omega) h + \sigma(t, \omega) B_h
\end{pmatrix},
\end{equation}
where $B_h\sim N(0, h \bm{I}_d)$ is a $d$-dimensional Gaussian vector.
For any vectorized matrix $y = ( x_0, x_1 , \ldots , x_i )^{\rm V} \in \real^{(i+1)d}$
with $0 \leq i \leq N$,
we consider the linear interpolation $\overline{y} \in \Omega$
 of $y$ defined as
\begin{equation}
\label{eq:interpolation}
 \overline{y}_s=
    \begin{cases}
      \displaystyle
      \frac{s-kh}{h} x_{k+1} + \left(1 - \frac{s-kh}{h} \right) x_k,
                & s \in [kh, (k+1)h),\ k =0, 1, \dots, i-1,
      \medskip
      \\
        x_i, & s \in [ih, T].
    \end{cases}
\end{equation}
For $\phi:\Theta \to \real$ a given function,
we let
\[
\mathcal{D}_h \phi(t, \omega) :=
\mathbb{E} \left[ \phi\big(t+h, \omega \otimes_{t} \overline{X}^{(t,\omega)}_h\big) H_h(t, \omega) \ \! \Big| \ \! \mathcal{F}_t \right],
\]
 where $H_h = \left( H_0^h, H_1^h, H_2^h \right)$ are the weights defined by
\[
	H_0^h:=1, \quad H_1^h:=\frac{B_h}{h}, \quad
	H_2^h:=\frac{B_h B^\top_h - h\bm{I}_d}{h^2}.
\]
As in (2.5) of \cite{fahim2011probabilistic}
and (4.11) of \cite{ren2017convergence},
we let the operator $\mathbb{T}^{t,\omega}$ be defined as
\begin{equation}\label{eq:operator_T}
\mathbb{T}^{t,\omega}\left[u^h(t+h, \ \! \cdot \ \! )\right]
	:= \mathbb{E} \left[ u^h\big(t+h, \omega \otimes_t \overline{X}^{(t,\omega)}_h\big) \ \! \Big| \ \! \mathcal{F}_t \right]
		+ hF\left(\ \! \cdot \ \! , \mathcal{D}_h u^h_{t+h}\right)(t,\omega).
\end{equation}
 The approximation $u^h$ of $u$ is then defined
 inductively as in (2.4) of \cite{fahim2011probabilistic}
 and \S~3 of \cite{ren2017convergence} as the linear interpolation
 $u^h(t, \omega)$ of the sequence
\begin{equation}\label{eq:probabilistic_scheme}
\begin{cases}
u^h(Nh, \omega) = g(\omega),
\medskip
\\
u^h( ih , \omega) = \mathbb{T}^{t,\omega}\left[u^h((i+1)h, \ \! \cdot \ \! )\right],
\qquad i = 0,1,\ldots , N-1.
\end{cases}
\end{equation}
The convergence of $u^h$ to $u$ as $h$ tends to zero
is ensured by the following result,
see Theorem~3.9 and Proposition~4.9
in \cite{ren2017convergence}.
\begin{theorem}
\label{theo:uh_to_u}
	Under Assumption~\ref{basic assumptions},
	assume further that the PPDE \eqref{eq:ppde}
        satisfies the comparison principle for
        viscosity subsolutions and supersolutions,
	i.e. if $v$ and $w$ are respectively
	viscosity subsolution and supersolution
	of PPDE \eqref{eq:ppde} and
	$v(T, \ \! \cdot \ \! ) \leq w(T, \ \! \cdot \ \! )$,
	then $v \leq w$ on $\Theta$.
	Then, the PPDE \eqref{eq:ppde} admits
	a unique viscosity solution $u$ given by the
        limit
	\begin{equation}
          \label{fjdksl}
	u(t,\omega ) = \lim_{h \rightarrow 0}
        u^h(t,\omega ),
\end{equation}
        locally uniformly in
        $(t,\omega )\in \Theta$.
        \end{theorem}
We refer to Theorem~4.2 of \cite{ren-touzi-zhang} for sufficient conditions
on PPDE coefficients for the 
comparison principle of viscosity solutions to be satisfied,
see also Section~\ref{fjdslk}. 
\noindent
 Convergence rates of $O(h^{1/10})$ and $O(h)$
 have been derived for \eqref{fjdksl} respectively in \cite{fahim2011probabilistic}
 for PDEs of Hamilton-Jacobi-Bellman type
 and in \cite{zhang2014monotone}
 for PPDEs under smoothness conditions,
 while the convergence rate of PPDE solutions
 remains unknown without smoothness conditions.

\section{Deep learning approximation}
\label{sec:deep}

\medskip

 In \cite{ren2017convergence},
the numerical estimation of the conditional expectation
in \eqref{eq:operator_T} has been implemented using regression
as in \cite{gobet2005regression},
which requires to choose a basis for the functional space
to be projected on.
 For example, assuming that 
 $$
 F(t, \omega, u, z, \gamma) = \widetilde{F}
 \left(t, \omega_t, \int_0^t \omega_s ds, u, z, \gamma \right)
 $$
 if the function $g$ in \eqref{eq:ppde}
 takes the form 
$$ 
 \quad g(\omega) = \widetilde{g}\left(\omega_T, \int_0^T \omega_s ds\right),
 $$
 it can be reasonably guessed that the actual solution $u$ will be of the form
 $$
 u(t, \omega) = \widetilde{u}\left(t, \omega_t, \int_0^t \omega_s ds\right),
 $$
motivating the choice of basis
$$
\left(1, \omega_t, \int_0^t \omega_s ds,
\omega^2_t, \left(\int_0^t \omega_s ds\right)^2, \omega_t\int_0^t \omega_s ds\right),
$$
 using second order polynomials.
 However, when $g$ is not expressed in such form, e.g.
 when
 $$
 \quad g(\omega) = \widetilde{g}\left(\omega_T, \sup_{s \in [0,T]} \omega_s\right),
 $$
it is less clear how the actual solution $u$ will look like,
making it difficult to pick an appropriate
basis for the projection.
Here, we overcome this difficulty 
by an alternative deep learning approach 
that does not rely on the specific form of
the actual solution $u$ and
has been previously applied with success
to various high-dimensional problems,
see e.g. \cite{han2018solving}, \cite{beck2019deep}, \cite{hure2019some}.

\medskip

Given $\rho:\real \to \real$ denote an activation function
such as $\rho_{\rm ReLU}(x) := \max (0,x)$,
$\rho_{\tanh}(x) := \tanh(x)$,
$\rho_{\rm Id}(x) := x$,
 we define the set of layer functions $\mathbb{L}^\rho_{d_1,d_2}$ by
\begin{equation}
\nonumber 
    \mathbb{L}^\rho_{d_1,d_2} :=
    \bigl\{
        L:\real^{d_1} \to \real^{d_2} \ : \ L(x) = \rho(Wx + b),
  \ x \in \real^{d_1}, \ W \in \real^{d_2 \times d_1}, \ b \in \real^{d_2}
    \bigr\},
\end{equation}
where $d_1 \geq 1$ is the input dimension,
$d_2 \geq 1$ is the output dimension,
and the activation function
$\rho$ is applied component-wise to $Wx + b$.
Then, we denote by
\begin{equation*}
    \mathbb{NN}^{\rho,l,m}_{d_0, d_1} :=
    \bigl\{
    L_l \circ \dots \circ L_0 : \real^{d_0} \to \real^{d_1}
    \ : \ 
        L_0 \in \mathbb{L}^{\rho}_{d_0, m},
        L_l \in \mathbb{L}^{\rho_{\rm Id}}_{m, d_1},
        L_i \in \mathbb{L}^\rho_{m, m},
        1 \leq i < l
    \bigr\}
\end{equation*}
the set of feed-forward neural networks
with one output layer,
$l \geq 1$ hidden layers
each containing $m \geq 1$ neurons,
and the activation functions of
the output layer and
the hidden layers
being respectively
the identity function $\rho_{\rm Id}$
and $\rho$.
 Any $L_l \circ \dots \circ L_0 \in \mathbb{NN}^{\rho,l,m}_{d_0, d_1}$
is fully determined by the sequence
\begin{equation*}
    \theta := \bigl( W_0, b_0, W_1, b_1, \dots, W_{l-1}, b_{l-1}, W_l, b_l \bigr),
\end{equation*}
 of $\left( (d_0+1) m + (l - 1) (m+1) m + (m+1) d_1 \right)$
 of parameters, such that
 $$        L_i (x) = \rho(W_l x + b_l),
 \qquad i=0,1,\ldots , l.
 $$
 Building on \eqref{eq:sde_discrete}-\eqref{eq:interpolation},
we let $X^\pi$ denote the discretization
\begin{equation}
    \label{eq:discretized_sde_ml}
    X^\pi_0 = x_0,
	\qquad
    X^\pi_{i+1} =
    \begin{pmatrix}
        X^\pi_i     \\
        X^{(ih, \overline{X}^\pi_i )}_h(1) - x_0 + X^\pi_i(i)
    \end{pmatrix},
    \quad
	i = 0, 1, \ldots, N-1,
\end{equation}
where $X^\pi_i (k) \in \real^d$ is the
 $k$-$th$ entry of the zero-based array $X^\pi_i \in \real^{(i+1)d}$
for $0 \leq k \leq i \leq N$.
The similar notation is also used on
$X^{(ih, \overline{X}^\pi_i )}_h$.
Next, we introduce the deep learning scheme
for the approximation of \eqref{eq:probabilistic_scheme}.
\begin{algo}
\label{algo:deep_ppde}
\begin{enumerate}[i)]
    \item Fix
        $(d,N,l,(m_i)_{0 \leq i < N })$,
        the activation function $\rho$,
        and a threshold $\varepsilon_{\rm thres}>0$,
        initialize $\widehat{\cal V}_N:\real^{(N+1)d} \to \real$
        by $\widehat{\cal V}_N(x)$ $=$ $g(\overline{x})$.
      \item For $i$ $=$ $N-1,\ldots,0$, given
          $\widehat{\cal V}_{i+1}:\real^{(i+2)d} \to \real$,
        \begin{enumerate}[a)]
          \item initialize the neural networks \\
              $\left({\cal Y}_i(\cdot \ ;\theta),
              {\cal Z}_i(\cdot \ ;\theta),
	      	  {\cal \gamma}_i(\cdot \ ;\theta)\right)
              \in \mathbb{NN}^{\rho,l,m_i}_{(i+1)d, 1}
              \times \mathbb{NN}^{\rho,l,m_i}_{(i+1)d, d}
              \times \mathbb{NN}^{\rho,l,m_i}_{(i+1)d, d(d+1)/2}$, 
            \item
              compute the mean square error function 
	      \begin{eqnarray} 
                              \nonumber
                E_i(\theta) & := & \E \bigl[ \big\lvert \widehat{\cal V}_{i+1}\big(X^\pi_{i+1}\big) H_0^h - {\cal Y}_i\big(X^\pi_i ;\theta\big)\big\rvert^2
			      + \big\lVert \widehat{\cal V}_{i+1}\big(X^\pi_{i+1}\big)H_1^h - {\cal Z}_i\big(X^\pi_i ;\theta\big)\big\rVert_d^2
                              \\
                \label{eq:deep_scheme_nonlinear}
                              &  & 
          + \big\lVert \widehat{\cal V}_{i+1}\big(X^\pi_{i+1}\big)H_2^h
      - {\rm Sym} \big( {\cal \gamma}_i\big(X^\pi_i ;\theta\big)\big) \big\rVert_{d\times d}^2\bigr],
              \end{eqnarray}
	      where
  	      for any sequence
              $(a_1, \dots, a_{d(d+1)/2}) \in \real^{d(d+1)/2}$ we let 
          \begin{align*}
              {\rm Sym} \big( (a_1, \dots, a_{d(d+1)/2})^\top \big)
                = &
                \begin{pmatrix}
                    2 a_{d(d-1)/2+1}  & a_{d(d+1)/2-1} & \dots        & a_2    & a_1    \\
                    a_{d(d+1)/2-1}    & \ddots  & \ddots     & \ddots & a_3    \\
                    \vdots          & \ddots        & \ddots       & \ddots & \vdots \\
                    a_2          & \ddots        & \ddots       & \ddots & a_{d(d-1)/2} \\
                    a_1               & a_3   & \dots          & a_{d(d-1)/2} & 2 a_{d(d+1)/2}
                \end{pmatrix}, 
          \end{align*}
	    \item
              choose
              $\theta_i^*$ in the set
              $$\left\{
              \theta = \bigl( W_0, b_0, W_1, b_1, \dots, W_{l-1}, b_{l-1}, W_l, b_l \bigr)
              \ : \
              E_i(\theta) <
                \inf\limits_{\theta} E_i(\theta) + \varepsilon_{\rm thres} \right\}. 
         	        $$
        \end{enumerate}
      \item Update
	      $\big(\widehat{\cal Y}_i( \cdot ),\widehat{\cal Z}_i( \cdot ),
	      		\widehat{\cal \gamma}_i( \cdot )\big)$
	      = $\left({\cal Y}_i(\cdot \ ;\theta_i^*),{\cal Z}_i(\cdot \ ;\theta_i^*),
	      		{\cal \gamma}_i(\cdot \ ;\theta_i^*)\right)$ and
          $\widehat{\cal V}_i:\real^{(i+1)d} \to \real$ by
          \begin{equation}\label{eq:define_v_hat}
            \widehat{\cal V}_i(x) := \widehat{\cal Y}_i(x) + h F\big( ih,
                \overline{x},
                \widehat{\cal Y}_i(x),
                \widehat{\cal Z}_i(x),
            {\rm Sym} \left( \widehat{\cal \gamma}_i(x) \right)\big).
          \end{equation}
\end{enumerate}
\end{algo}
 We note that minimizing the error function $E_i(\theta)$
is equivalent to minimizing the quantity 
\begin{eqnarray}
  \nonumber
  \varepsilon_i^{{l,m}, \theta}
 & := &
\E \big[ \big\lvert\E_i \big[\widehat{\cal V}_{i+1}\left(X^\pi_{i+1}\right) H_0^h\big] - {\cal Y}_i\left(X^\pi_i ;\theta\right)\big\rvert^2
		    + \big\lVert\E_i\big[\widehat{\cal V}_{i+1}\left(X^\pi_{i+1}\right)H_1^h\big] - {\cal Z}_i\left(X^\pi_i ;\theta\right)\big\rVert_d^2
                    \\
                    \label{fdsk}
	            & &
                    + \big\lVert\E_i\big[\widehat{\cal V}_{i+1}\left(X^\pi_{i+1}\right)H_2^h\big] - {\rm Sym} \left({\cal \gamma}_i\left(X^\pi_i ;\theta\right)\right) \big\rVert_{d\times d}^2\big], 
\end{eqnarray}
 from the relationship
\begin{align}
		    \nonumber
 E_i(\theta) & = \E \big[ \big\lvert \widehat{\cal V}_{i+1}\big(X^\pi_{i+1}\big) H_0^h - \E_i\big[\widehat{\cal V}_{i+1}\big(X^\pi_{i+1}\big) H_0^h\big]
			+ \E_i\big[\widehat{\cal V}_{i+1}\big(X^\pi_{i+1}\big) H_0^h\big] - {\cal Y}_i\big(X^\pi_i ;\theta\big)\big\rvert^2
		    \\ \nonumber
		    & \quad + \big\lVert \widehat{\cal V}_{i+1}\big(X^\pi_{i+1}\big)H_1^h - \E_i\big[\widehat{\cal V}_{i+1}\big(X^\pi_{i+1}\big)H_1^h\big]
			    + \E_i\big[\widehat{\cal V}_{i+1}\big(X^\pi_{i+1}\big)H_1^h\big] - {\cal Z}_i\big(X^\pi_i ;\theta\big)\big\rVert_d^2
		    \\ \nonumber
		    & \quad + \big\lVert \widehat{\cal V}_{i+1}\big(X^\pi_{i+1}\big)H_2^h - \E_i\big[\widehat{\cal V}_{i+1}\big(X^\pi_{i+1}\big)H_2^h\big]
        + \E_i\big[\widehat{\cal V}_{i+1}\big(X^\pi_{i+1}\big)H_2^h\big] - {\rm Sym} \big( {\cal \gamma}_i\big(X^\pi_i ;\theta\big)\big) \big\rVert_{d\times d}^2\big]
		    		    \\ \nonumber
		    & = \E \big[ \big\lvert \widehat{\cal V}_{i+1}\left(X^\pi_{i+1}\right) H_0^h - \E_i\big[\widehat{\cal V}_{i+1}\left(X^\pi_{i+1}\right) H_0^h\big]\big\rvert^2
 + \big\lVert \widehat{\cal V}_{i+1}\left(X^\pi_{i+1}\right)H_1^h - \E_i\big[\widehat{\cal V}_{i+1}\left(X^\pi_{i+1}\right)H_1^h\big]\big\rVert_d^2
 \\
 \nonumber 
		    & \quad + \big\lVert \widehat{\cal V}_{i+1}\left(X^\pi_{i+1}\right)H_2^h - \E_i\big[\widehat{\cal V}_{i+1}\left(X^\pi_{i+1}\right)H_2^h\big]\big\rVert_{d\times d}^2
	    \big] + \varepsilon_i^{{l,m}, \theta},
\end{align}
where in the second equality, we have used the fact that
for any square-integrable $\mathcal{F}_i$-measurable random variable $Y$,
\[
\E [ (\E_i [X] -Y) (X - \E_i[X])] = \E [(\E_i[X] -Y)\E_i [ X - \E_i[X] ] ] = 0.
\]
 The next result is an error bound of the Algorithm~\ref{algo:deep_ppde}.
  \begin{prop}
   \label{theo:deep_scheme_nonlinear}
  Using \eqref{eq:probabilistic_scheme} and the notation of
   Algorithm~\ref{algo:deep_ppde},
  and assuming
 $F (t, \omega, u, z, \gamma)$ is
	  Lipschitz with respect to $(\omega, u, z, \gamma)
          \in \Omega \times \real \times \real^d \times \mathbb{S}^d$
	  uniformly in $t \in [0,T]$
          as in
  part~\eqref{aaa} of Assumption~\ref{basic assumptions},
   we have
 \begin{align}
   \nonumber 
    \max_{i=0,\ldots,N-1} \E \big[ \big|\widehat{\cal V}_{i}\left(X^\pi_i \right) - u^h\big(ih, \overline{X}^\pi_{ih}\big)\big|^2 \big] \leq M \frac{L^N - 1}{L-1}\varepsilon^{l ,m},
\end{align}
 where we let $L := 32\left(1+K^2h^2 + K^2hd + K^2 d(d+1) \right)$,
 $M := 32\big(1 + K^2h^2\big)$, and 
\begin{equation}
  \label{gfjkfld}
\varepsilon^{l,m} := \sum\limits_{i = 0}^{N-1} \varepsilon_i^{{l,m}, \theta_i^*}. 
\end{equation}
\end{prop}
\begin{Proof}
  Let $\delta_i := \widehat{\cal V}_{i}\left(X^\pi_i \right) - u^h\big(ih, \overline{X}^\pi_{ih}\big)
  $,
  $i=0,\ldots , N-1$.
By \eqref{eq:probabilistic_scheme}, \eqref{eq:define_v_hat},
Assumption~\ref{basic assumptions} and the conditional H{\"o}lder inequality, 
we have
\begin{align*}
&	\E \left[|\delta_i|^2\right] = \E \big[ \bigl\lvert \widehat{\cal Y}_i\left(X^\pi_i \right)
        + h F\big(ih, \overline{X}^\pi_{ih}, \widehat{\cal Y}_i\left(X^\pi_i\right),
        \widehat{\cal Z}_i\left(X^\pi_i\right),{\rm Sym} \left(\widehat{\cal \gamma}_i\left(X^\pi_i\right)\right) \big)
            - \E_i\left[ u^h\big((i+1)h, \overline{X}^\pi_{(i+1)h}\big)\right]  \\
                             & \quad + hF\bigl(ih, \overline{X}^\pi_{ih}, \E_i\left[ u^h\big((i+1)h, \overline{X}^\pi_{(i+1)h}\big)H_0^h\right], \E_i\left[ u^h\big((i+1)h, \overline{X}^\pi_{(i+1)h}\big)H_1^h\right], \\
                             & \qquad\qquad \E_i\left[ u^h\big((i+1)h, \overline{X}^\pi_{(i+1)h}H_2^h\big)\right]\bigr) \bigr\rvert^2\big] \\
		    & \leq 16\big(1 + K^2h^2\big) \E
		    	\big[\big\lvert \widehat{\cal Y}_i\left(X^\pi_i\right)
                - \E_i\left[ u^h\big((i+1)h, \overline{X}^\pi_{(i+1)h}\big)H_0^h\right]\big\rvert^2 \big]  \\
		    & \quad + 16K^2h^2 \E \big[\big\lVert \widehat{\cal Z}_i\left(X^\pi_i\right)
                - \E_i\left[ u^h\big((i+1)h, \overline{X}^\pi_{(i+1)h}\big)H_1^h\right]\big\rVert_d^2  \\
            & \quad \qquad\qquad\qquad + \big\lVert {\rm Sym} \left( \widehat{\cal \gamma}_i\left(X^\pi_i\right) \right)
        -  \E_i\left[ u^h\big((i+1)h, \overline{X}^\pi_{(i+1)h}\big)H_2^h\right]\big\rVert_{d\times d}^2 \big] \\
		    & = 16\big(1 + K^2h^2\big) \E \big[\bigl\lvert \widehat{\cal Y}_i\left(X^\pi_i\right) - \E_i\big[ \widehat{\cal V}_{i+1}\big(X^\pi_{i+1}\big)H_0^h\big] \\
		    & \quad\qquad\qquad + \E_i\big[ \widehat{\cal V}_{i+1}\left(X^\pi_{i+1}\right)H_0^h\big]
    - \E_i\left[ u^h\big((i+1)h, \overline{X}^\pi_{(i+1)h}\big)H_0^h\right]\bigr\rvert^2 \big] \\
		        & \quad + 16K^2h^2 \E \big[ \bigl\lVert \widehat{\cal Z}_i\left(X^\pi_i\right) - \E_i\big[ \widehat{\cal V}_{i+1}\left(X^\pi_{i+1}\right)H_1^h\big]
                          + \E_i\big[ \widehat{\cal V}_{i+1}\left(X^\pi_{i+1}\right)H_1^h\big]
                          \\
                          & \quad\qquad\qquad
                          -  \E_i\left[ u^h\big((i+1)h, \overline{X}^\pi_{(i+1)h}\big)H_1^h\right]\bigr\rVert_d^2  + \bigl\lVert {\rm Sym} \left( \widehat{\cal \gamma}_i\left(X^\pi_i\right) \right) - \E_i\big[ \widehat{\cal V}_{i+1}\left(X^\pi_{i+1}\right)H_2^h\big] \\
            & \quad\qquad\qquad + \E_i\big[ \widehat{\cal V}_{i+1}\left(X^\pi_{i+1}\right)H_2^h\big] -  \E_i\big[ u^h\big((i+1)h, \overline{X}^\pi_{(i+1)h}\big)H_2^h\big]\bigr\rVert_{d\times d}^2 \big] \\
          & \leq 32\big(1 + K^2h^2\big) \varepsilon_i^{{l ,m}, \theta^*}
	  + 32\E \big[ \E_i\left[\delta_{i+1}\right]\E_i\big[(1+K^2h^2)\left\lvert H_0^h \right\rvert^2
		  + K^2h^2\left\lVert H_1^h \right\rVert_d^2 + K^2h^2\left\lVert H_2^h \right\rVert_{d\times d}^2\big] \big] \\
          & \leq 32\big(1 + K^2h^2\big) \varepsilon_i^{{l ,m}, \theta^*}
          + 32\left(1+K^2h^2 + K^2hd + K^2 d(d+1) \right) \E\left[ \delta_{i+1} \right]
	  \\
           & = M\varepsilon_i^{{l ,m}, \theta^*} + L \E\left[ \delta_{i+1} \right],
\end{align*}
$i=0,\ldots , N-2$.
By backward induction and
the fact that $\widehat{\cal V}_{N}(x) = u^h(t_N, \overline{x}) = g(\overline{x})$,
we obtain
\begin{align*}
\max_{i=0,\ldots,N-1} \E [ |\delta_i|^2 ] &\leq \sum\limits_{i = 0}^{N-1} L^{i-1}M \varepsilon_i^{{l ,m}, \theta^*} \leq M \varepsilon^{l ,m} \sum\limits_{i = 0}^{N-1} L^{i-1} = M \frac{L^N - 1}{L-1}\varepsilon^{l ,m}.
\end{align*}
\end{Proof}
The proof of Proposition~\ref{theo:deep_scheme_nonlinear}
uses only the Lipschitz continuity of $F$ in Assumption~\ref{basic assumptions},
while the rest of the conditions in Assumption~\ref{basic assumptions}
are required in Theorem~\ref{theo:uh_to_u}
as in \cite{fahim2011probabilistic} and \cite{ren2017convergence}.
 Next, we recall the following universal approximation theorem.
\begin{theorem}
  \label{t1}
  (Theorem~1 in \cite{hornik1991approximation}).
    Fix $l  \ge 1$, 
    if the activation function $\rho$ is unbounded and nonconstant, then
    for any finite measure $\mu$ the set
    $\bigcup\limits_{m = 1}^{\infty} \mathbb{NN}^{\rho,l ,m}_{d_0, 1}$
    is dense in $L^q(\mu)$ for all $q \ge 1$.
\end{theorem}
The next corollary shows that the neural network approximation
can be made arbitrarily close to the PPDE solution $u\left(0, (x_0)_{s \in [0,T]}\right)$.
\begin{theorem}
    \label{corollary_convergence}
  Under the assumptions of Theorems~\ref{theo:uh_to_u} and \ref{t1},
  assume additionally that the activation function
  $\rho$ is Lipschitz.
  Then, for any $\varepsilon >0$ there exists
  $(m_i)_{0 \leq i < N }$
  and $(\theta_i^*)_{0 \leq i < N }$ such that
  $(\widehat{\cal V}_i)_{0 \leq i \leq N}$ constructed from
  $(m_i)_{0 \leq i < N }$ and $(\theta_i^*)_{0 \leq i < N }$
  in \eqref{eq:define_v_hat} satisfies
\begin{equation*}
	\big\lvert u (0, (x_0)_{s \in [0,T]} )
                - \widehat{\cal V}_{0}(x_0)\big\rvert < \varepsilon.
\end{equation*}
\end{theorem}
\begin{Proof}
Let $\varepsilon >0$. By Theorem~\ref{theo:uh_to_u},
we can find $h > 0$ small enough such that
\begin{equation}
\nonumber 
    \left\lvert u\left(0, (x_0)_{s \in [0,T]}\right)
                     - u^h\left(0, (x_0)_{s \in [0,T]}\right)\right\rvert < \frac{\varepsilon}{2}.
\end{equation}
First, we note that by Proposition~\ref{theo:deep_scheme_nonlinear}, the proof is complete
by the triangle inequality if we can choose
$(m_i)_{0 \leq i < N }$ and $(\theta_i^*)_{0 \leq i < N }$
 such that $\varepsilon^{l,m}$ defined by \eqref{fdsk} and \eqref{gfjkfld} satisfies
 \begin{align}
    \label{small_epsilon}
\nonumber
\varepsilon^{l,m}
& = \sum\limits_{i = 0}^{N-1} \E\Bigl[ \big\lvert\E_i\big[
        \widehat{\cal V}_{i+1}\left(X^\pi_{i+1}\right) H_0^h\big]
                - {\cal Y}_i\left(X^\pi_i;\theta_i^*\right)\big\rvert^2
 + \big\lVert\E_i\big[
        \widehat{\cal V}_{i+1}\left(X^\pi_{i+1}\right)H_1^h\big]
                - {\cal Z}_i\left(X^\pi_i;\theta_i^*\right)\big\rVert_d^2
\\
& \quad + \big\lVert\E_i\big[
        \widehat{\cal V}_{i+1}\left(X^\pi_{i+1}\right)H_2^h\big]
    - {\rm Sym} \left({\cal \gamma}_i\left(X^\pi_i;\theta_i^*\right)\right)
    \big\rVert_{d\times d}^2 \Bigr] \ < \
 \frac{L-1}{2M(L^N - 1)}
 \varepsilon,
\end{align}
Next, we note that \eqref{small_epsilon} holds if we show that
\begin{equation}
\nonumber 
    \E \big[ \big\lVert\E_i\big[
        \widehat{\cal V}_{i+1}\big(X^\pi_{i+1}\big)H_2^h\big]
    - {\rm Sym} \big({\cal \gamma}_i\big(X^\pi_i;\theta_i^*\big)\big)
                \big\rVert_{d\times d}^2 \big] <  \frac{L-1}{6NM(L^N - 1)}\varepsilon,
\end{equation}
 $i=0,\ldots , N-1$, as the argument for the other terms similar.
 For this, we rely on
the universal approximation Theorem~\ref{t1},
which requires us to show that 
$\E_i\big[\widehat{\cal V}_{i+1}\big(X^\pi_{i+1}\big)H_2^h\big]$,
as a function of $X_i$ on $\real^{(i+1)d} \to \mathbb{S}^d$, 
is in $L^2(\mu)$, where $\mu$ is the joint distribution of
 $X^\pi_i$. 
By the Lipschitz condition on $b (t, \omega)$ and $\sigma (t, \omega)$
in Assumption~\ref{basic assumptions} we have
\begin{equation}
\label{regularity_Xpi}
\E \Big[\max\limits_{0 \leq k \leq i} \norm{X^\pi_i(k)}_d^q\Big] < \infty
\qquad \text{and} \qquad
\E \left[\norm{\varphi(X^\pi_i)}_k^q\right] < \infty,
\end{equation}
for any Lipschitz continuous function $\varphi:\real^{(i+1)d} \to \real^k$
and all $q \ge 1$, $0 \leq i \leq N$, see Appendix~\ref{appendix}.
    Hence, by the Lipschitz Assumption~\ref{basic assumptions}-\eqref{part2}
 and H\"older's inequality, we have
$\E_i\big[\widehat{\cal V}_{i+1}\big(X^\pi_{i+1}\big)H_2^h\big]
= \E_i\big[g\big( \overline{X}^\pi_{(i+1)h}\big)H_2^h\big]
\in L^2(\mu)$ at the level $i = N-1$.
For $0 \leq i < N$, using Assumption~\ref{basic assumptions}-\eqref{aaa} we have
\begin{align*}
    & \E \big[ \big\lVert \E_i\big[\widehat{\cal V}_{i+1}\big(X^\pi_{i+1}\big)
        H_2^h\big]\big\rVert_{d\times d}^2 \big]
    \\
      &\leq \ \E \bigl[  \norm{H_2^h}_{d\times d}^2 \times
                \bigl\lvert \widehat{\cal Y}_{i+1}(X^\pi_{i+1})
                    + h F\big( (i+1)h, \overline{X}^\pi_{(i+1)h},
                \widehat{\cal Y}_{i+1}(X^\pi_{i+1}),
    \\
      &\qquad \qquad \qquad \qquad \qquad \qquad \qquad
                \widehat{\cal Z}_{i+1}(X^\pi_{i+1}),
                {\rm Sym} \left( \widehat{\cal \gamma}_{i+1}(X^\pi_{i+1})
                \right)\big) \bigr\rvert^2
                        \bigr]
    \\
      &\leq \ 6^2 \Bigl( \E \bigl[\norm{H_2^h}_{d\times d}^4 \bigr]
      \E\bigl[  \abs{\widehat{\cal Y}_{i+1}(X^\pi_{i+1})}^4
          + h^4 \abs{F((i+1)h, (0)_{0 \leq s \leq T}, 0, 0,0)}^4
    \\
      &\quad\quad
      + h^4 K^4 \bigl(\norm{\overline{X}^\pi_{(i+1)h}}^4
          +\abs{\widehat{\cal Y}_{i+1}(X^\pi_{i+1})}^4
          +\norm{\widehat{\cal Z}_{i+1}(X^\pi_{i+1})}_d^4
          +\norm{{\rm Sym} \big( \widehat{\cal \gamma}_{i+1}(X^\pi_{i+1})
      \big)}_{d\times d}^4\bigr) \bigr] \Bigr)^{1/2}
    \\
      &< \ \infty,
\end{align*}
which shows that
$\E_i\big[\widehat{\cal V}_{i+1}\big(X^\pi_{i+1}\big)H_2^h\big]\in L^2(\mu)$.
In the last inequality we have used
\eqref{regularity_Xpi}, the fact that
$\varphi \in \mathbb{NN}^{\rho,l ,m}_{d_0, d_1}$
is Lipschitz when the activation function $\rho$ is Lipschitz,
 and $\E \left[ \norm{H^h_2}_{d\times d}^4 \right] < \infty$
 with $\norm{\overline{X}^\pi_{ih}} : = \max\limits_{0 \leq k \leq i} \norm{X^\pi_i(k)}_d$.
\end{Proof}
Using additionally that
$u \in $ BUC$(\Theta)$ is uniformly continuous,
 Proposition~\ref{theo:deep_scheme_nonlinear}
 and Theorem~\ref{corollary_convergence}
 can be extended from $(0, (x_0)_{s \in [0,T]})$ to any $(t, \omega) \in \Theta$
 by changing \eqref{eq:discretized_sde_ml}
to start from $X^\pi_{kh} = (\omega^\pi_{s \land kh})_{s \in [0,T]}$,
where $\omega^\pi$ is the linear interpolation of the
 discretization of $\omega$.

\section{Numerical examples}
\label{sec:numerical}
The optimization in
\eqref{eq:deep_scheme_nonlinear}
is implemented using Monte Carlo simulation
and the Adam gradient descent algorithm,
see \cite{kingma2014adam}.
Precisely, fix the batch size $O$ and the training steps $P$,
let $\big(X^{\pi,j}_{(i+1)}\big)_{1 \leq j \leq O}$
be an i.i.d. sample of $(i+1)d$-dimensional random vector
with batch size $O$ generated by \eqref{eq:discretized_sde_ml}, and let
\begin{align*}
    L^O_i(\theta) &:= \frac{1}{O}\sum\limits_{j=1}^O \bigl[ \big\lvert \widehat{\cal V}_{i+1}\big(X^{\pi,j}_{i+1}\big) H_0^h - {\cal Y}_i\big(X^{\pi,j}_i;\theta\big)\big\rvert^2
        + \big\lVert \widehat{\cal V}_{i+1}\big(X^{\pi,j}_{i+1}\big)H_1^h - {\cal Z}_i\big(X^{\pi,j}_{i};\theta\big)\big\rVert_d^2
                  \\
                  & \qquad + \big\lVert \widehat{\cal V}_{i+1}\big(X^{\pi,j}_{i+1}\big)H_2^h
              - {\rm Sym} \big( {\cal \gamma}_i\big(X^{\pi,j}_i ;\theta\big)\big) \big\rVert_{d\times d}^2\bigr].
\end{align*}
Then, we initialize the parameter $\theta_0$
using Xavier initialization, see \cite{glorot2010understanding},
and update it using the following rule
\begin{align*}
  \begin{cases}
    \displaystyle
    v_p         &
    \displaystyle
    = \beta_1 v_{p-1} + (1-\beta_1)
  \frac{\partial L^O_i}{\partial \theta} (\theta_{p-1}) \\
\displaystyle
w_p         &
\displaystyle
    = \beta_2 w_{p-1} + (1-\beta_2) \left(
  \frac{\partial L^O_i}{\partial \theta} (\theta_{p-1})\right)^2 \\
\displaystyle
\theta_p    &
\displaystyle
    = \theta_{p-1} - \eta_p \left( \frac{v_p}{1-\beta_1} \right)
    \bigg/ \left( \varepsilon_{\rm Adam} + \sqrt{\frac{w_p}{1-\beta_1}} \right),
  \end{cases}
\end{align*}
where $1 \leq p \leq P$, $(\eta_p)_{1 \leq p \leq P} \in \real^P$ is the learning rate,
$(\varepsilon_{\rm Adam},\beta_1,\beta_2) \in \real^3$ are the parameters
 of the Adam algorithm,
and $(v_0, w_0)$ is initialized at $(0,0)$.
Empirically we have $\theta_P \approx \theta^*$
when $O$ and $P$ are large enough, see e.g. \cite{kingma2014adam}.
    In addition, we use the batch normalization technique, see \cite{ioffe2015batch},
    to stabilize the training process.
    Define $BN_{\gamma,\beta,\varepsilon_{BN}}$ a transformation over
    a set of $d_1$-dimensional $\big(x^{(i)}_j\big)_{1 \leq i \leq O, 1 \leq j \leq d_1}$
    with batch size $O$
    by
    \begin{equation}
        BN_{\gamma,\beta,\varepsilon_{BN}}\big(x^{(i)}\big) =
        \Big( \beta_j + \gamma_j \big(x^{(i)}_j - \mu_j\big) \big/
            \sqrt{\sigma_j^2 + \varepsilon_{BN}} \ \Big)_{1 \leq j \leq d_1} ,
    \end{equation}
    where
    $\gamma , \beta \in \real^{d_1}$, $\varepsilon_{BN} \in \real$,
    and
    $$
    \mu_j = \frac{1}{O} \sum\limits_{i=1}^O x^{(i)}_j
    \quad \mbox{and} \quad
    \sigma_j^2 = \frac{1}{O} \sum\limits_{i=1}^O (x^{(i)}_j - \mu_j)^2.
    $$
    Fix $\varepsilon_{BN} \in \real$, a neural network
    $\varphi( \cdot \ ; \theta ) \in \mathbb{NN}^{\rho,l ,m}_{d_0, d_1}$
    is modified such that each of the layer functions
    $L_i \in \mathbb{L}^\rho_{d_2,d_3}$ is changed to
    $L_i \in \mathbb{L}^{\rho,BN}_{d_2,d_3}$, where
    \begin{equation*}
        \mathbb{L}^{\rho,BN}_{d_2,d_3} :=
        \Bigl\{
            L:\real^{d_2} \to \real^{d_3} \ : \ L(x) =
            BN_{\gamma,\beta,\varepsilon_{BN}}\left( \rho(W x + b) \right),
            W \in \real^{d_3 \times d_2},
            b, \gamma, \beta \in \real^{d_3}
        \Bigr\},
    \end{equation*}
    and a transformation from
    $x \in \real^{d_0}$ to $BN_{\gamma,\beta,\varepsilon_{BN}}(x)$
    is added before passing to the first layer.
    Then, the neural network parameter $\theta$ is changed to
    \begin{equation*}
        \theta^{BN} = \bigl( \gamma_{-1}, \beta_{-1}W_0, b_0, \gamma_0, \beta_0,
            W_1, b_1, \gamma_1, \beta_1, \dots,
            W_{l -1}, b_{l -1}, \gamma_{l -1}, \beta_{l -1},
            W_l , b_l , \gamma_l , \beta_l  \bigr).
    \end{equation*}
In the following subsections, we provide
three examples of implementation of
the numerical scheme of Proposition~\ref{theo:deep_scheme_nonlinear}.
In our numerical examples we use the activation function $\rho = \rho_{\rm ReLU}$,
the Adam parameters $\left(\beta_1, \beta_2,\varepsilon_{\rm Adam}\right)
= \left(0.9, 0.999,10^{-8}\right)$,
the batch normalization parameter
$\varepsilon_{BN} = 10^{-6}$,
and the learning rate
\begin{equation*}
    \eta_p =
    \begin{cases}
        10^{-1},   &1    \leq p < 2P/3,
        \\
        10^{-2},   &2P/3 \leq p < 5P/6,
        \\
        10^{-3},   &5P/6 \leq p < P. 
    \end{cases}
\end{equation*}
The numerical simulations of \cite{saporito2020pdgm} and \cite{sabate2020solving} are not presented in dimension $d=100$ because they require more than the 12 GB RAM provided by Google Colab.

\medskip

 As in \cite{alanko2013reducing},
 for better convergence we implement the modification
\begin{equation*}
        \begin{cases}
        Y_h\phi(t, \omega) := \mathbb{E} \left[ \phi\left(t+h,
          \omega \otimes_{t} \overline{X}^{(t,\omega)}_h\right) H^h_0
          \ \! \Big| \ \! \mathcal{F}_t \right],
        \medskip \\
        Z_h\phi(t, \omega) := \mathbb{E} \left[ \left( \phi\left(t+h,
                        \omega \otimes_{t} \overline{X}^{(t,\omega)}_h\right)  -
                        Y_h\phi(t, \omega) \right) H^h_1
                        \ \! \Big| \ \! \mathcal{F}_t
                        \right],
        \medskip \\
        \Gamma_h\phi(t, \omega) := \mathbb{E} \left[ \left( \phi\left(t+h,
                        \omega \otimes_{t} \overline{X}^{(t,\omega)}_h\right)  - Y_h\phi(t, \omega)
                        - Z_h\phi(t, \omega) \cdot W_h \right) H^h_2
                        \ \! \Big| \ \! \mathcal{F}_t
                        \right],
        \end{cases}
    \end{equation*}
of \eqref{eq:probabilistic_scheme}
    where the operator $\mathbb{T}^{t,\omega}$
    in \eqref{eq:operator_T}
    is replaced by
    \begin{equation}
    \label{eq:var_reduced_probabilistic_scheme}
    \mathbb{T}^{t,\omega}\left[u^h(t+h, \ \! \cdot \ \! )\right]
        = Y_h\phi(t, \omega) + hF\left(\ \! \cdot \ \! , Y_h\phi, Z_h\phi, \Gamma_h\phi \right)(t,\omega),
    \end{equation}
    and the error function $E_i(\theta)$
    in \eqref{eq:deep_scheme_nonlinear} is replaced with
    \begin{align}
	\nonumber
        E_i(\theta) & =  \E \bigl[ \big\lvert \widehat{\cal V}_{i+1}\big(X^\pi_{i+1}\big) H_0^h - {\cal Y}_i\big(X^\pi_i;\theta\big)\big\rvert^2
            + \big\lVert \big( \widehat{\cal V}_{i+1}\big(X^\pi_{i+1}\big) - {\cal Y}_i\big(X^\pi_i;\theta\big) \big)H_1^h - {\cal Z}_i\big(X^\pi_i;\theta\big)\big\rVert_d^2
    \\
    \label{eq:var_reduced_deep_scheme_nonlinear}
            & \quad + \big\lVert \big( \widehat{\cal V}_{i+1}\big(X^\pi_{i+1}\big)
                        - {\cal Y}_i\big(X^\pi_i;\theta\big)
                        - {\cal Z}_i\big(X^\pi_i;\theta\big) \cdot W_h \big) H_2^h
                    - {\cal \gamma}_i\big(X^\pi_i;\theta\big)\big\rVert_{d\times d}^2\bigr].
    \end{align}
  Although the following examples do not satisfy all 
  conditions stated in Assumption~\ref{basic assumptions},
  we will use them as in e.g. \cite{ren2017convergence}
  to assess the performance of  Algorithm~\ref{algo:deep_ppde}. 
   In the sequel, we let
    $(B_t)_{0 \leq t \leq T} = (B^1_t, \dots, B^d_t)_{0 \leq t \leq T}$
  denote a $d$-dimensional Brownian motion.
\subsection{Path-dependent two-person zero-sum game}
\label{fjdslk} 
  In this section we consider the higher dimensional extension
\begin{equation}
\nonumber 
    u(0, x_0) = \inf\limits_{\mu_t \in [\ubar{\mu}, \overline{\mu}]}
            \sup\limits_{a_t \in [\ubar{a}, \overline{a}]}
            \E \left[ g\left(X^{\mu, a}_T, \int_0^T X^{\mu, a}_s ds\right)
                + \int_0^T f\left(t, X^{\mu, a}_t, \int_0^t X^{\mu, a}_s ds\right) dt \right],
\end{equation}
of the path-dependent two-person zero-sum game
in (5.1) of \cite{ren2017convergence},
 where $(X_t)_{0 \leq t \leq T} = (X^1_t, \dots, X^d_t)_{0 \leq t \leq T}$
follows the SDE
\begin{equation}
\begin{cases}
    dX^{\mu, a}_t = \mu_t \bm{1}_d dt + \sqrt{a_t} \ \bm{I}_d dB_t , \\
    X_0 = x_0,
\end{cases}
\end{equation}
where $x_0 \in \real^d$,
$\ubar{\mu}$, $\overline{\mu}$, $\ubar{a}$, $\overline{a} \in \real$.
 The solution of this control problem is the solution of the following PPDE
$u:[0,T] \times C ([0,T]; \real^d ) \to \real$
evaluated at $(0, x_0)$
\begin{equation*}
  \partial_t u 
  + \min\limits_{\mu \in [\ubar{\mu}, \overline{\mu}]} \mu \big(
  \bm{1}_d \cdot \partial_\omega u\big)
  +
  \frac{1}{2}
  \max\limits_{a \in [\ubar{a}, \overline{a}]}
  \left( a 
    \tr\left(\partial^2_{\omega\omega}u\right)\right)
    + f\left(t, \omega_t, \int_0^t \omega_s ds\right)
    = 0, \quad
    u(T, \omega) = g\left(\omega_T, \int_0^T \omega_s ds\right),
\end{equation*}
and for the purpose of our deep algorithm
we rewrite the above PPDE as
\begin{equation}\label{eq:ppde_control}
  \partial_t u + \frac{\ubar{a}}{2} \tr \left(\partial^2_{\omega\omega}u\right)
    +     \min\limits_{\mu \in [\ubar{\mu}, \overline{\mu}]} \mu \left(\bm{1}_d \cdot \partial_\omega u\right)
    + \frac{1}{2} \max\limits_{a \in [\ubar{a}, \overline{a}]} a 
    \tr\left(\partial^2_{\omega\omega}u\right)
    + f\left(t, \omega_t, \int_0^t \omega_s ds\right)
    - \frac{\ubar{a}}{2} \tr\left(\partial^2_{\omega\omega}u\right)
        = 0,
\end{equation}
with
$$
F (t, \omega, u, z, \gamma)
= 
\frac{1}{\sqrt{\ubar{a}}}
\min\limits_{\mu \in [\ubar{\mu}, \overline{\mu}]} \mu \left(\bm{1}_d \cdot
 z  \right)
+ \frac{1}{2
\ubar{a}
} \max\limits_{a \in [\ubar{a}, \overline{a}]}
\big( a \tr \gamma \big) 
    + f\left(t, \omega_t, \int_0^t \omega_s ds\right)
    - \frac{1}{2} \tr {\gamma}
    .
    $$
     Denoting by $x = (x^1, \ldots, x^d)$ and $y = (y^1, \ldots, y^d)$
we put $g(x, y) = \cos\left(\frac{1}{d}\sum\limits_{i = 1}^d \left(x^i+y^i\right)\right)$ and
\begin{align*}
    f(t, x, y) = & \left(\frac{1}{d}\sum\limits_{i = 1}^d x^i + \overline{\mu}\right)\left(\sin\left(\frac{1}{d}\sum\limits_{i = 1}^d \left(x^i+y^i\right)\right)\right)^{+}
                  -\left(\frac{1}{d}\sum\limits_{i = 1}^d x^i + \ubar{\mu}\right)\left(\sin\left(\frac{1}{d}\sum\limits_{i = 1}^d \left(x^i+y^i\right)\right)\right)^{-}
                 \\
                 & +\frac{\ubar{a}}{2d}\left(\cos\left(\frac{1}{d}\sum\limits_{i = 1}^d \left(x^i+y^i\right)\right)\right)^{+}
                -\frac{\overline{a}}{2d}\left(\cos\left(\frac{1}{d}\sum\limits_{i = 1}^d \left(x^i+y^i\right)\right)\right)^{-}.
\end{align*}
Although this choice of $f(t,x,y)$ does not satisfy  part~\eqref{aaa} of Assumption~\ref{basic assumptions},
it makes the PPDE \eqref{eq:ppde_control} explicitly solvable as
$\displaystyle
 u(t, \omega) = \cos \bigg(\frac{1}{d}\sum\limits_{i = 1}^d \left(\omega_t^i+\int_0^t \omega_s^i ds \right)\bigg)$,
 which can be used to evaluate the precision of
 Algorithm~\ref{algo:deep_ppde}.
 Source codes are available on request.
  
\begin{table}[H]
    \centering
		\resizebox{\textwidth}{!}{\begin{tabular}{|c|c|c|c|c|c|c|c|}
			\hline
                        Method & $d$ & Regr./Deep & Mean & Stdev & Ref. value & Rel. $L^1$-error & 
                        Runtime (s)\\
            \hline
            Deep PPDE using \eqref{eq:var_reduced_deep_scheme_nonlinear} & 1 & Deep & 1.000805 & 9.61E-05 & 1.0 & 8.05E-04 & 62 \\
            Deep PPDE using \eqref{eq:deep_scheme_nonlinear} & 1 & Deep & 0.999331 & 1.52E-03 & 1.0 & 1.37E-03 & 61 \\
            \cite{saporito2020pdgm} & 1 & Deep & 1.000852 & 1.12E-02 & 1.0 & 9.42E-03 & 26 \\
                     \cline{3-8}
            \cite{ren2017convergence} using \eqref{eq:var_reduced_probabilistic_scheme} & 1 & Regr. & 0.9999463 & 4.72E-05 & 1.0 & 5.47E-05 & 1 \\
            \cite{ren2017convergence} using \eqref{eq:probabilistic_scheme} & 1 & Regr. & 1.075509 & 2.59E-02 & 1.0 & 7.55E-02 & 1 \\
            \hline
            Deep PPDE using \eqref{eq:var_reduced_deep_scheme_nonlinear} & 10 & Deep & 1.000914 & 2.19E-04 & 1.0 & 9.14E-04 & 63 \\
            Deep PPDE using \eqref{eq:deep_scheme_nonlinear} & 10 & Deep & 0.9939934 & 2.99E-03 & 1.0 & 6.01E-03 & 62 \\
            \cite{saporito2020pdgm} & 10 & Deep & 0.9537241 & 1.63E-01 & 1.0 & 1.06E-01 & 517 \\
                 \cline{3-8}
                 \cite{ren2017convergence} using \eqref{eq:var_reduced_probabilistic_scheme} & 10 & Regr. & 1.000166 & 6.00E-06 & 1.0 & 1.66E-04 & 2
                 \\
            \cite{ren2017convergence} using \eqref{eq:probabilistic_scheme} & 10 & Regr. & 2.348812 & 4.31E-01 & 1.0 & Diverges & 2 \\
            \hline
            Deep PPDE using \eqref{eq:var_reduced_deep_scheme_nonlinear} & 100 & Deep & 1.002474 & 5.01E-04 & 1.0 & 2.47E-03 & 83 \\
            Deep PPDE using \eqref{eq:deep_scheme_nonlinear} & 100 & Deep & 0.970783 & 1.89E-02 & 1.0 & 3.01E-02 & 81 \\
            \cline{3-8}
            \cite{ren2017convergence} using \eqref{eq:var_reduced_probabilistic_scheme} & 100 & Regr. & 26.12039 & 7.36E+00 & 1.0 & Diverges & 104 \\
            \cite{ren2017convergence} using \eqref{eq:probabilistic_scheme} & 100 & Regr. & 328.2853 & 8.81E+01 & 1.0 & Diverges & 102 \\
            \hline
		\end{tabular}}
		\caption{
            Comparison between 
            \ $i)$ PPDE with training parameters
                $m = d+10$, $l  = 2$, $O = 256$, $h = 0.01$,
                and $P = 900$;
            \ $ii)$ \cite{ren2017convergence} with
                $O = 10000$ and $h = 0.01$;
            \ $iii)$ \cite{saporito2020pdgm} with training parameters
                $m = d+10$, $l  = 2$, $O = 256$, $h = 0.01$,
                and $P = 1000$.}
        \label{table:control_highdimension}
\end{table}
\noindent
 In Table~\ref{table:control_highdimension}, our PPDE algorithm is compared to
\cite{ren2017convergence} and \cite{saporito2020pdgm}
 with ten Monte Carlo runs,
 with $\ubar{\mu} = -0.2$, $\overline{\mu} = 0.2$, $\protect \ubar{a} = 0.04$, $\overline{a} = 0.09$, $T = 0.1$, $x_0 = (0, \ldots, 0)$,
 and runtimes are measured in seconds.

\subsection{Asian options}
\label{jklds}
The second example is the following pricing problem of Asian basket call option:
\begin{equation}
\nonumber 
    u(0, x_0) = \E \left[ e^{-r_0T} \left(\frac{1}{Td}\sum\limits_{i = 1}^d  \int_0^T X^i_s ds - K \right)^+ \right],
\end{equation}
with strike price $K \in \real$ and interest rate $r_0>0$, where
 $(X_t)_{0 \leq t \leq T} = (X^1_t, \dots, X^d_t)_{0 \leq t \leq T}$
is a $d$-dimensional asset price process following
the geometric Brownian motions
\begin{equation}
  \label{gbm}
X_t^i = X_0^i \re^{\sigma_iB_t^i + r_i -\sigma_i^2t/2}, \qquad
t\in \real_+, \quad i = 1,\ldots , d,
\end{equation}
where $x_0 \in \real^d$,
and $r_1, \ldots, r_d, \sigma_1, \ldots, \sigma_d \in \real$.
The solution of this pricing problem is given by
evaluating at $(t,x)=(0, x_0)$
the solution $u:[0,T] \times C\left([0,T]; \real^d\right) \rightarrow \real$
 of the following PPDE:
\begin{equation}
\nonumber 
    \partial_t u + r(\omega_t) \cdot \partial_\omega u
    + \frac{1}{2} \left( \sigma\sigma^\top  (\omega_t)  : \partial^2_{\omega\omega}u \right) - r_0u = 0,
    \ \ u(T, \omega) = \left( \frac{1}{Td}\sum\limits_{i = 1}^d  \int_0^T \omega^i_s ds - K \right)^+,
\end{equation}
 where $r(\omega_t) = \left(r_1\omega^1_t, \ldots, r_d\omega^d_t\right)$ and
 $\sigma(\omega_t) = \diag \left(\sigma_1\omega^1_t, \ldots, \sigma_d\omega^d_t\right)$.
 Here, $F(t, \omega, u, z, \gamma) = -r_0 u$ does not depend on
 $z$ and $\gamma$, therefore
the neural networks ${\cal Z}_i(\cdot \ ;\theta)$ and
${\cal \gamma}_i(\cdot \ ;\theta)$
 are not needed, 
 which improves the efficiency of Algorithm~\ref{algo:deep_ppde}.

\medskip

 When $d=1$ we compare our deep PPDE algorithm 
 with other deep PDE algorithms such as
 \cite{han2018solving} and \cite{beck2019deep}.
 For this, we write
 $$
 u(t, (X_s)_{s \in [0,t]} ) = g\left(t, \frac{1}{X_t} \left(\frac{1}{T} \int_0^t X_u dt - K\right)\right),
 $$
 where $g:[0,T] \times \real \rightarrow \real$ is the solution of
 the \cite{rogersshi} PDE
 \begin{equation}
   \partial_t g +  (1/T - rz) \frac{\partial g}{\partial x}
    + \frac{1}{2} \sigma^2 z^2 \partial^2_{zz}g = 0, \quad g(T, z) = z^+,
\end{equation}
 see e.g. Proposition~13.10 in \cite{privaultbkf2}.

 \medskip

We use Monte Carlo simulations
with $O=1,000,000$ and $h=0.01$
as the reference solution.
to compare our PPDE algorithm with
\cite{ren2017convergence}, \cite{saporito2020pdgm},
\cite{sabate2020solving}, \cite{han2018solving}, and \cite{beck2019deep}
under the setting of
$r_0 = r_1 = \cdots = r_d = 0.01$,
$\sigma_1 = \cdots = \sigma_d = 0.1$,
$K = 0.7$, $T = 0.1$, $x_0 = (1, \ldots, 1)$.
The statistics of 10 independent runs
are summarized in Table~\ref{table:asian_highdimension}.
\begin{table}[H]
    \centering
		\resizebox{\textwidth}{!}{\begin{tabular}{|c|c|c|c|c|c|c|c|}
			\hline
                        Method & $d$ & Regr./Deep & Mean & Stdev & Ref. value & Rel. $L^1$-error & 
                        Runtime (s)\\
            \hline
            \cite{han2018solving} & 1 & Deep & 0.3002467 & 2.31E-06 & 0.3002021 & 1.49E-04 & 32 \\
            \cite{beck2019deep} & 1 & Deep & 0.3002827 & 4.21E-04 & 0.3002021 & 1.12E-03 & 20 \\
            \cite{sabate2020solving} & 1 & Deep & 0.3002722 & 1.24E-03 & 0.3002021 & 3.51E-03 & 10 \\
            Deep PPDE & 1 & Deep & 0.3008159 & 1.29E-03 & 0.3002021 & 3.83E-03 & 31 \\
            \cite{saporito2020pdgm} & 1 & Deep & 0.3002544 & 2.43E-03 & 0.3002021 & 6.01E-03 & 25 \\
            \cline{3-8}
            \cite{ren2017convergence} & 1 & Regr. & 0.3002768 & 2.23E-04 & 0.3002021 & 4.77E-04 & 1 \\
            \hline
            Deep PPDE & 10 & Deep & 0.3010345 & 4.08E-04 & 0.3002024 & 2.77E-03 & 31 \\
            \cite{sabate2020solving} & 10 & Deep & 0.3002251 & 2.11E-03 & 0.3002024 & 5.80E-03 & 411 \\
            \cite{saporito2020pdgm} & 10 & Deep & 0.304033 & 1.05E-02 & 0.3002024 & 2.97E-02 & 522 \\
            \cline{3-8}
            \cite{ren2017convergence} & 10 & Regr. & 0.3002137 & 6.07E-05 & 0.3002024 & 1.70E-04 & 2 \\
            \hline
            Deep PPDE & 100 & Deep & 0.3006346 & 1.28E-04 & 0.3001993 & 1.45E-03 & 35 \\
            \cline{3-8}
            \cite{ren2017convergence} & 100 & Regr. & 0.3001923 & 2.65E-05 & 0.3001993 & 7.50E-05 & 23 \\
            \hline
		\end{tabular}}
		\caption{
            Comparison between 
            \ $i)$ PPDE with training parameters
                $m = d+10$, $l  = 2$, $O = 256$, $h = 0.01$,
                and $P = 900$;
            \ $ii)$ \cite{ren2017convergence} with
                $O = 10000$ and $h = 0.01$;
            \ $iii)$ \cite{saporito2020pdgm} with training parameters
                $m = d+10$, $l  = 2$, $O = 256$, $h = 0.01$,
                and iterations 1000;
            \ $iv)$ \cite{sabate2020solving} with training parameters
                $m = d+10$, $l  = 2$, $O = 256$, $h = 0.01$,
                and $P = 600$;
            \ $v)$ \cite{han2018solving} with training parameters
                $m = d+10$, $l  = 2$, $O = 64$, $h = 0.01$,
                and $P = 4000$;
            \ $vi)$ \cite{beck2019deep} with training parameters
                $m = d+10$, $l  = 2$, $O = 256$, $h = 0.01$,
                and $P = 600$.}
        \label{table:asian_highdimension}
\end{table}

\subsection{Barrier options}
The third example is the following pricing problem of barrier basket call option:
\begin{equation}
\nonumber 
    u(0, x_0) = \E \left[ e^{-r_0T} \mathbbm{1}_{\Big\{ \max\limits_{0 \leq s \leq T} \Big(\frac{1}{d}\sum\limits_{i = 1}^d X^i_s\Big) < B \Big\}}\left(\frac{1}{d}\sum\limits_{i = 1}^d X^i_T - K \right)^+ \right],
\end{equation}
where the strike price $K \in \real$,
the barrier $B \in \real$,
and $(X_t)_{0 \leq t \leq T} = (X^1_t, \dots, X^d_t)_{0 \leq t \leq T}$
is a $d$-dimensional stock processes that follows
the geometric Brownian motions \eqref{gbm}
The solution of this pricing problem is given by
evaluating at $(t,x)=(0, x_0)$
the solution $u:[0,T] \times C\left([0,T]; \real^d\right) \rightarrow \real$
 of the following PPDE:
\begin{align}
    \nonumber
    & \partial_t u + r(\omega_t) \cdot \partial_\omega u
    + \frac{1}{2} \left( \sigma\sigma^\top (\omega_t):\partial^2_{\omega\omega}u \right) - r_0u = 0,
    \\
\nonumber 
    & u(T, \omega) = \mathbbm{1}_{\Big\{ \max\limits_{0 \leq s \leq T} \Big(\frac{1}{d}\sum\limits_{i = 1}^d \omega^i_s\Big) < B \Big\}}\left(\frac{1}{d}\sum\limits_{i = 1}^d \omega^i_T - K \right)^+.
\end{align}
As in Section~\ref{jklds},
 $F(t, \omega, u, z, \gamma) = -r_0 u$ does not depend on
 $z$ and $\gamma$ and
the neural networks ${\cal Z}_i(\cdot \ ;\theta)$, and
${\cal \gamma}_i(\cdot \ ;\theta)$
are not needed.

\medskip 

We use Monte Carlo simulations
with $O=1000000$ and $h=0.01$
as the reference solution
to compare our PPDE algorithm with
\cite{ren2017convergence}, \cite{saporito2020pdgm}, and \cite{sabate2020solving}
under the setting of
$r_0 = r_1 = \cdots = r_d = 0.01$, $\sigma_1 = \cdots = \sigma_d = 0.1$,
$K = 0.7$, $B = 1.2$, $T = 0.1$, $x_0 = (1, \ldots, 1)$.
The statistics of 10 independent runs
are summarized in Table~\ref{table:barrier_highdimension}.
\begin{table}[H]
    \centering
		\resizebox{\textwidth}{!}{\begin{tabular}{|c|c|c|c|c|c|c|c|}
			\hline
                        Method & $d$ & Regr./Deep & Mean & Stdev & Ref. value & Rel. $L^1$-error & 
                        Runtime (s)\\
            \hline
            \cite{sabate2020solving} & 1 & Deep & 0.3009402 & 2.18E-03 & 0.3007008 & 5.75E-03 & 8 \\
            Deep PPDE & 1 & Deep & 0.3019161 & 1.97E-03 & 0.3007008 & 5.92E-03 & 31 \\
            \cite{saporito2020pdgm} & 1 & Deep & 0.3019159 & 2.24E-03 & 0.3007008 & 6.98E-03 & 26 \\
            \cline{3-8}
            \cite{ren2017convergence} & 1 & Regr. & 0.3006738 & 2.81E-04 & 0.3007008 & 7.72E-04 & 1 \\
            \hline
            Deep PPDE & 10 & Deep & 0.3017532 & 5.44E-04 & 0.3006973 & 3.51E-03 & 31 \\
            \cite{sabate2020solving} & 10 & Deep & 0.301107 & 3.15E-03 & 0.3006973 & 7.35E-03 & 225 \\
            \cite{saporito2020pdgm} & 10 & Deep & 0.3030515 & 1.03E-02 & 0.3006973 & 2.71E-02 & 519 \\
            \cline{3-8}
            \cite{ren2017convergence} & 10 & Regr. & 0.3007255 & 1.34E-04 & 0.3006973 & 3.56E-04 & 2 \\
            \hline
            Deep PPDE & 100 & Deep & 0.3016375 & 1.98E-04 & 0.3007003 & 3.12E-03 & 35 \\
            \cline{3-8}
            \cite{ren2017convergence} & 100 & Regr. & 0.3035602 & 3.74E-03 & 0.3007003 & 1.05E-02 & 23 \\
            \hline
\end{tabular}}
		\caption{
            Comparison between 
            \ $i)$ PPDE with training parameters
                $m = d+10$, $l  = 2$, $O = 256$, $h = 0.01$,
                and $P = 900$;
            \ $ii)$ \cite{ren2017convergence} with
                $O = 10000$ and $h = 0.01$;
            \ $iii)$ \cite{saporito2020pdgm} with training parameters
                $m = d+10$, $l  = 2$, $O = 256$, $h = 0.01$,
                and $P = 1000$;
            \ $iv)$ \cite{sabate2020solving} with training parameters
                $m = d+10$, $l  = 2$, $O = 256$, $h = 0.01$,
                and $P = 600$.}
        \label{table:barrier_highdimension}
\end{table}

\appendix

\section{Appendix}
\label{appendix}
\noindent
Proof of \eqref{regularity_Xpi}.
 We first show the finiteness of
the first term in \eqref{regularity_Xpi} by induction,
where $i=0$ obviously holds.
Assume that it holds at level $i$, by Assumption~\ref{basic assumptions},
H\"older's inequality, the independence between $B_h$ and $\overline{X}^\pi_{ih}$,
and \eqref{eq:discretized_sde_ml}, we have
\begin{eqnarray*}
  \lefteqn{
    \E \Big[\max\limits_{0 \leq k \leq i+1} \norm{X^\pi_{i+1}(k)}_d^q\Big]
            \leq
                \E \Big[\max\Big( \max\limits_{0 \leq k \leq i} \norm{X^\pi_i(k)}_d^q,
                \norm{X^\pi_{i+1}((i+1)h)}_d^q \Big)\Big]
  }
  \\
            &= &
                \E \Big[\max\Big( \max\limits_{0 \leq k \leq i} \norm{X^\pi_i(k)}_d^q,
                \ \norm{X^\pi_i(i)
                        + b\big(ih, \overline{X}^\pi_{ih}\big)h
                        + \sigma\big(ih, \overline{X}^\pi_{ih}\big)B_h }_d^q \Big)\Big]
            \\
            &\leq &
            \E \Big[ \max\limits_{0 \leq k \leq i} \norm{X^\pi_i(k)}_d^q\Big] +
                \E \left[ \norm{X^\pi_i(i)
                        + b\big(ih, \overline{X}^\pi_{ih}\big) h
                        + \sigma\big(ih, \overline{X}^\pi_{ih}\big)B_h }_d^q\right]
            \\
            &\leq &
            \E \Big[ \max\limits_{0 \leq k \leq i} \norm{X^\pi_i(k)}_d^q\Big] +
                3^q \E \Bigl[ \norm{X^\pi_i(i)}_d^q
  + h^q \big(K \big(\abs{ih}^{1/2} + \norm{\overline{X}^\pi_{ih}}\big) +
                    \norm{b(0,(0)_{0\leq s\leq T})}_d \big)^q
            \\
            & &  \quad + \norm{B_h}_d^q \big(K \big(\abs{ih}^{1/2} + \norm{\overline{X}^\pi_{ih}}\big) +
                \norm{\sigma(0,(0)_{0\leq s\leq T})}_{d\times d} \big)^q\Bigr]
            \\
            &\leq &
            \E \Big[ \max\limits_{0 \leq k \leq i} \norm{X^\pi_i(k)}_d^q\Big] +
                9^q \E \left[ \norm{X^\pi_i(i)}_d^q \right]
  + h^q \left(K^q \left(T^{q/2} + \E \left[ \norm{\overline{X}^\pi_{ih}}^q\right]\right) +
                    \norm{b(0,(0)_{0\leq s\leq T})}_d^q \right)
            \\
            & &  \quad + \E \left[ \norm{B_h}_d^q \right] \left(K^q \left(T^{q/2} + \E \left[ \norm{\overline{X}^\pi_{ih}}^q\right]\right) +
                    \norm{\sigma(0,(0)_{0\leq s\leq T})}_{d\times d}^q \right)
            \\
            &\leq & C_0 + C_1 \E \Big[\max\limits_{0 \leq k \leq i}
                    \norm{X^\pi_i(k)}_d^q\Big] < \infty,
\end{eqnarray*}
where $C_0 \ge 0$ and $C_1 \ge 1$.
In the second last inequality we used the fact that
$\norm{\overline{X}^\pi_{ih}}^q = \max\limits_{0 \leq k \leq i} \norm{X^\pi_i(k)}_d^q$,
and the centered Gaussian random variable $B_h$
has finite $\E \left[ \abs{B_h}^q \right]$ for any choice of $q$.
The finiteness of the second term in \eqref{regularity_Xpi}
is obvious since
\begin{equation*}
    \E \left[\norm{\varphi(X^\pi_i)}_k^q\right] \le
                K^q \E \big[\norm{\overline{X}^\pi_{ih}}_{(i+1)d}^q \big] +
                    \norm{\varphi(0,(0)_{0\leq s\leq T})}_k^q
        < \infty.
\end{equation*}

\footnotesize

\setcitestyle{numbers}

\def\cprime{$'$} \def\polhk#1{\setbox0=\hbox{#1}{\ooalign{\hidewidth
  \lower1.5ex\hbox{`}\hidewidth\crcr\unhbox0}}}
  \def\polhk#1{\setbox0=\hbox{#1}{\ooalign{\hidewidth
  \lower1.5ex\hbox{`}\hidewidth\crcr\unhbox0}}} \def\cprime{$'$}

\section{Python code}
\label{sec:sourcecode}

\begin{lstlisting}[language=Python,
caption={\it deep\_ppde.py},
label={code:ppde}
]
import tensorflow as tf
import numpy as np
import time


########## Section neural network ####################
class FullModel(tf.keras.Model):
    """
    Full Model that contains the list of all neural networks
    and the initial y0, z0, g0
    """
    def __init__(self, config, eqn):
        super(FullModel, self).__init__()
        self.config = config
        self.eqn = eqn
        self.subnet = [FeedForwardSubNet(config, eqn)
                      for _ in range(config.N-1)]
        y0 = tf.Variable(tf.random_uniform_initializer()(
                          shape=[1,1], dtype=config.dtype))
        if self.eqn.ppde_type == 'linear':
            self.subnet.append([y0])
        elif self.eqn.ppde_type == 'semilinear':
            z0 = tf.Variable(tf.random_uniform_initializer()(
                          shape=[1,config.dim], dtype=config.dtype))
            self.subnet.append([y0, z0])
        elif self.eqn.ppde_type == 'fully_nonlinear':
            z0 = tf.Variable(tf.random_uniform_initializer()(
                          shape=[1,config.dim], dtype=config.dtype))
            g0 = tf.Variable(tf.random_uniform_initializer()(
                          shape=[1,config.dim**2], dtype=config.dtype))
            self.subnet.append([y0, z0, g0])

    def __call__(self, x, training, nn_type, idx):
        if nn_type=='y':
            if idx==(self.config.N-1):
                return self.subnet[idx][0]
            else:
                return self.subnet[idx](x, training, 0)
        elif nn_type=='z':
            if idx==(self.config.N-1):
                return self.subnet[idx][1]
            else:
                return self.subnet[idx](x, training, 1)
        elif nn_type=='g':
            if idx==(self.config.N-1):
                return self.subnet[idx][2]
            else:
                return self.subnet[idx](x, training, 2)

    def build(self, idx):
        """
        Make sure the model is run at least once
        before passing to the graph generated by tf.function
        """
        # the last model is not NN, hence no initialization is needed
        if idx==(self.config.N-1): return

        x = tf.zeros([1, self.config.dim*(self.config.N-idx)])
        self.subnet[idx](x, False, 0)

        if self.eqn.ppde_type == 'semilinear' or \
              self.eqn.ppde_type == 'fully_nonlinear':
            self.subnet[idx](x, False, 1)

        if self.eqn.ppde_type == 'fully_nonlinear':
            self.subnet[idx](x, False, 2)


class FeedForwardSubNet(tf.keras.Model):
    """Implementation of individual neural networks."""
    def __init__(self, config, eqn):
        super(FeedForwardSubNet, self).__init__()
        self.config = config
        self.eqn = eqn
        self.bn_layers = [[tf.keras.layers \
                             .BatchNormalization(dtype=config.dtype)
                           for _ in range(len(config.y_neurons)+1)]]
        self.dense_layers = [[tf.keras.layers.Dense(
                                                config.y_neurons[i],
                                                dtype=config.dtype,
                                                use_bias=False,
                                                activation=None)
                              for i in range(len(config.y_neurons))]]

        # Z network for the semilinear or fully nonlinear PPDE
        if self.eqn.ppde_type == 'semilinear' \
                or self.eqn.ppde_type == 'fully_nonlinear':
            self.bn_layers.append([
                tf.keras.layers.BatchNormalization(dtype=config.dtype)
                    for _ in range(len(config.z_neurons) + 1)])
            self.dense_layers.append([
                tf.keras.layers.Dense(config.z_neurons[i],
                                      dtype=config.dtype,
                                      use_bias=False,
                                      activation=None)
                    for i in range(len(config.z_neurons))])

        # Gamma network for the fully nonlinear PPDE
        if self.eqn.ppde_type == 'fully_nonlinear':
            self.bn_layers.append([
                tf.keras.layers.BatchNormalization(dtype=config.dtype)
                    for _ in range(len(config.g_neurons) + 1)])
            self.dense_layers.append([
                tf.keras.layers.Dense(config.g_neurons[i],
                                      dtype=config.dtype,
                                      use_bias=False,
                                      activation=None)
                    for i in range(len(config.g_neurons))])

    def __call__(self, x, training, nn_type):
        """
        bn -> (dense->bn->relu)*(depth-1) -> dense -> bn
        """
        x = tf.reshape(x, [x.shape[0], -1])
        x = self.bn_layers[nn_type][0](x, training)
        depth = len(self.dense_layers[nn_type])
        for i in range(depth):
            x = self.dense_layers[nn_type][i](x)
            x = self.bn_layers[nn_type][i+1](x, training)
            if i < (depth-1):
                x = tf.nn.relu(x)
        return x


########## Section equation ####################
class Equation:
    """Base class for defining the problem."""
    def __init__(self, config):
        self.config = config
        self.x_init = None

    def b(self, x):
        """Drift of the forward SDE."""
        raise NotImplementedError

    def sigma(self, x):
        """Diffusion of the forward SDE."""
        raise NotImplementedError

    def sigma_inverse(self, x):
        """Inverse of the diffusion of the forward SDE."""
        # if this function is not implemented,
        # we use inv function in tf.linalg
        return tf.linalg.inv(self.sigma(x))

    def sde(self, n):
        """Simulate forward SDE."""
        x = [self.x_init*tf.ones((self.config.batch_size,
                self.config.dim), dtype=self.config.dtype)]
        for _ in range(n + 1):
            dw = tf.random.normal((self.config.batch_size,
                self.config.dim), stddev=self.config.sqrt_delta_t,
                dtype=self.config.dtype)
            # squeeze used because matmul produces dim of d*1
            # while the rest are d
            x.append(x[-1] + self.b(x[-1]) * self.config.delta_t
                    + tf.squeeze(tf.matmul(self.sigma(x[-1]),
                        tf.expand_dims(dw, -1)), -1))
        x = tf.stack(x, axis=1)
        return x, dw

    def f(self, x, y, z, gamma):
        """Generator of the PPDE."""
        raise NotImplementedError

    def phi(self, x):
        """Terminal condition of the PPDE."""
        raise NotImplementedError

    def symmetrise(self, gamma):
        # reshape to get the correct [d, d] dimension
        # then symmetrize the matrix using upper triangular part
        # because the Hessian matrix is a symmetric matrix
        gamma = tf.reshape(gamma, [-1,self.config.dim,self.config.dim])
        gamma = tf.linalg.band_part(gamma, 0, -1)
        gamma = 0.5 * (gamma + tf.transpose(gamma, perm=[0, 2, 1]))
        return gamma


class ControlProblem(Equation):
    """Example in Section 5.1"""
    def __init__(self, config):
        super(ControlProblem, self).__init__(config)
        self.ppde_type = 'fully_nonlinear'   # fully nonlinear PPDE
        self.x_init = 0
        self.mu_low = -0.2
        self.mu_high = 0.2
        self.sig_low = 0.2
        self.sig_high = 0.3
        self.a_low = self.sig_low ** 2
        self.a_high = self.sig_high ** 2

    def b(self, x):
        return tf.zeros_like(x)

    def sigma(self, x):
        return self.sig_low * tf.eye(self.config.dim,
                batch_shape=[self.config.batch_size])

    def sigma_inverse(self, x):
        return tf.eye(self.config.dim,
                  batch_shape=[self.config.batch_size]) / self.sig_low

    def f(self, x, y, z, gamma):
        # sum along the dimension axis
        reduced_x = tf.reduce_mean(x, 2)
        # trapezoidal rule
        reduced_int = self.config.delta_t * (tf.reduce_sum(reduced_x,1)
                        + tf.reduce_sum(reduced_x[:,1:-1], 1)) / 2
        reduced_sin = tf.sin(reduced_x[:,-1] + reduced_int)
        reduced_cos = tf.cos(reduced_x[:,-1] + reduced_int)
        reduced_z = tf.reduce_sum(z,1)
        reduced_gamma = tf.linalg.trace(gamma)
        cancel = -self.a_low * reduced_gamma / 2
        mu = tf.where(reduced_z>0, self.mu_low*reduced_z,
                self.mu_high*reduced_z)
        a = tf.where(reduced_gamma>0, self.a_high*reduced_gamma,
                self.a_low*reduced_gamma) / 2
        small_f = reduced_sin * tf.where(reduced_sin>0,
                                reduced_x[:,-1] + self.mu_high,
                                reduced_x[:,-1] + self.mu_low) \
                    + tf.where(reduced_cos>0,
                            self.a_low*reduced_cos/2,
                            self.a_high*reduced_cos/2) / self.config.dim
        return tf.expand_dims(cancel+mu+a+small_f, -1)

    def phi(self, x):
        # average along the dimension axis
        reduced_x = tf.reduce_mean(x, 2)
        # trapezoidal rule
        reduced_int = self.config.delta_t * (tf.reduce_sum(reduced_x,1)
                        + tf.reduce_sum(reduced_x[:,1:-1], 1)) / 2
        return tf.expand_dims(tf.cos(reduced_x[:, -1]
                + reduced_int), -1)


class AsianOption(Equation):
    """Example in Section 5.2"""
    def __init__(self, config):
        super(AsianOption, self).__init__(config)
        self.ppde_type = 'linear'   # linear PPDE
        self.x_init = 1
        self.sig = 0.1
        self.r = 0.01
        self.K = 0.7

    def b(self, x):
        return self.r*x

    def sigma(self, x):
        return self.sig * tf.linalg.diag(x)

    def sigma_inverse(self, x):
        return tf.linalg.diag(tf.reciprocal(x)) / self.sig

    def f(self, x, y, z, gamma):
        return -self.r*y

    def phi(self, x):
        # average along the dimension axis
        reduced_x = tf.reduce_mean(x, 2)
        # trapezoidal rule
        reduced_mean = self.config.delta_t*(tf.reduce_sum(reduced_x,1)\
                + tf.reduce_sum(reduced_x[:,1:-1],1))/(2*self.config.T)
        reduced_mean -= self.K
        return tf.expand_dims(tf.where(reduced_mean>0,
            reduced_mean, tf.zeros_like(reduced_mean)), -1)


class BarrierOption(Equation):
    """Example in Section 5.3"""
    def __init__(self, config):
        super(BarrierOption, self).__init__(config)
        self.ppde_type = 'linear'   # linear PPDE
        self.x_init = 1
        self.sig = 0.1
        self.r = 0.01
        self.K = 0.7
        self.B = 1.2

    def b(self, x):
        return self.r*x

    def sigma(self, x):
        return self.sig * tf.linalg.diag(x)

    def sigma_inverse(self, x):
        return tf.linalg.diag(tf.reciprocal(x)) / self.sig

    def f(self, x, y, z, gamma):
        return -self.r*y

    def phi(self, x):
        # average along the dimension axis
        reduced_x = tf.reduce_mean(x, 2)
        reduced_term = reduced_x[:, -1] - self.K
        up_flag = self.B - tf.reduce_max(reduced_x, 1)
        return tf.expand_dims(tf.where(up_flag>0, reduced_term,
            tf.zeros_like(reduced_term)), -1)


########## Section PPDE solver ####################
class PPDESolver:
    """
    Define the relationship between the variables of FullModel
    according to the type of Equation.
    """
    def __init__(self, config, eqn):
        self.config = config
        self.eqn = eqn
        self.model = FullModel(config, eqn)
        self.v0 = None
        # self.idx = None
        self.lr_schedule = tf.keras \
                             .optimizers \
                             .schedules \
                             .PiecewiseConstantDecay(
                                 config.lr_boundaries,config.lr_values)
        self.epsilon = 1e-8
        self.optimizer = None

    def train(self):
        for idx in range(self.config.N):
            # generate new graph using new idx with tf.function
            # this significantly speeds up the computation
            tf_train_step = tf.function(self.train_step)
            # self.idx = idx
            self.optimizer = tf.keras \
                               .optimizers \
                               .Adam(learning_rate=self.lr_schedule,
                                      epsilon=self.epsilon)
            self.model.build(idx)
            for step in range(self.config.train_steps):
                loss, v0 = tf_train_step(idx)

            del tf_train_step   # throw away old graph
        self.v0 = tf.reduce_mean(v0).numpy()   # v0 is 1x1 tensor

    def train_step(self, idx):
        x, dw = self.eqn.sde(self.config.N-idx-1)
        trainable_var = self.model.subnet.trainable_variables
        with tf.GradientTape(persistent=True) as tape:
        # with tf.GradientTape() as tape:
            loss, v0 = self.loss_fn(x, dw, idx, training=True)
        grads = tape.gradient(loss, trainable_var)
        del tape

        self.optimizer.apply_gradients(
                  (grad,var) for (grad,var) in zip(grads,trainable_var)
                  if grad is not None)
        return loss, v0

    def loss_fn(self, x, dw, idx, training):
        v0 = None   # the solution of interest defined in advance

        ### loss regarding y network ######
        if idx == 0:   # at terminal time
            y_target = self.eqn.phi(x)
        else:
            y_temp = self.model(x, False, 'y', idx-1)
            # fully nonlinear problem requires 3 networks
            if self.eqn.ppde_type == 'fully_nonlinear':
                z_temp = self.model(x, False, 'z', idx-1)
                g_temp = self.model(x, False, 'g', idx-1)
                g_temp = self.eqn.symmetrise(g_temp)
                y_target = y_temp + self.config.delta_t \
                            * self.eqn.f(x, y_temp, z_temp, g_temp)
            # semilinear problem requires 2 networks
            elif self.eqn.ppde_type == 'semilinear':
                z_temp = self.model(x, False, 'z', idx-1)
                y_target = y_temp + self.config.delta_t \
                            * self.eqn.f(x, y_temp, z_temp, z_temp)
            # linear problem requires only 1 networks
            else:
                y_target = y_temp + self.config.delta_t \
                            * self.eqn.f(x, y_temp, y_temp, y_temp)

        # the solution v0 should be fixed throughout the batch
        if idx == self.config.N-1:
            xnow = tf.slice(x, [0,0,0], [1,1,self.config.dim])
            y0 = self.model(xnow, training, 'y', idx)
            y_now = tf.tile(y0, [self.config.batch_size, 1])
            if self.eqn.ppde_type == 'linear':
                v0 = y0 + self.config.delta_t \
                            * self.eqn.f(xnow, y0, y0, y0)
        else:
            y_now = self.model(x[:, :-1, :], training, 'y', idx)

        loss = tf.reduce_mean((y_now-tf.stop_gradient(y_target)) ** 2)
        # we are done here for linear model
        if self.eqn.ppde_type == 'linear':
            return loss, v0

        ### loss regarding z network ######
        sig_inverse = self.eqn.sigma_inverse(x[:, -2, :])
        # expand_dims is needed for tensor multiplication
        # because y_target is 1d but dw is nd
        if self.config.var_reduction:
            # minus y_now for variance reduction
            z_target = tf.expand_dims(y_target - y_now, -1) \
                        * tf.matmul(sig_inverse,
                                tf.expand_dims(dw, -1),
                                transpose_a=True) \
                        / self.config.delta_t
        else:
            z_target = tf.expand_dims(y_target, -1) \
                        * tf.matmul(sig_inverse,
                                tf.expand_dims(dw, -1),
                                transpose_a=True) \
                        / self.config.delta_t

        # the solution v0 should be fixed throughout the batch
        if idx == self.config.N-1:
            xnow = tf.slice(x, [0,0,0], [1,1,self.config.dim])
            z0 = self.model(xnow, training, 'z', idx)
            z_now = tf.tile(z0, [self.config.batch_size, 1])
            if self.eqn.ppde_type == 'semilinear':
                v0 = y0 + self.config.delta_t \
                            * self.eqn.f(xnow, y0, z0, z0)
        else:
            z_now = self.model(x[:, :-1, :], training, 'z', idx)

        loss += tf.reduce_mean((z_now
                    - tf.stop_gradient(tf.squeeze(z_target,-1))) ** 2)
        # we are done here for semilinear model
        if self.eqn.ppde_type == 'semilinear':
            return loss, v0

        ### loss regarding g network ######
        dw2 = tf.matmul(tf.expand_dims(dw, -1),
                tf.expand_dims(dw, -1), transpose_b=True)
        if self.config.var_reduction:
            g_target = (tf.expand_dims(y_target-y_now,-1) \
                        - tf.matmul(
                            tf.matmul(self.eqn.sigma(x[:, -2, :]),
                              tf.expand_dims(z_now, -1),
                              transpose_a=True ),
                            tf.expand_dims(dw, -1),
                            transpose_a=True)) \
                    * tf.matmul(
                        tf.matmul(sig_inverse,
                          (dw2 - self.config.delta_t
                           * tf.eye(self.config.dim,
                              batch_shape=[self.config.batch_size])),
                           transpose_a=True),
                      sig_inverse) / (self.config.delta_t)**2
        else:
            g_target = tf.expand_dims(y_target,-1) \
                       * tf.matmul(
                         tf.matmul(sig_inverse,
                            (dw2 - self.config.delta_t
                             * tf.eye(self.config.dim,
                                batch_shape=[self.config.batch_size])),
                             transpose_a=True),
                         sig_inverse) / (self.config.delta_t)**2

        # the solution v0 should be fixed throughout the batch
        if idx == self.config.N-1:
            xnow = tf.slice(x, [0,0,0], [1,1,self.config.dim])
            g0 = self.model(xnow, training, 'g', idx)
            g_now = tf.tile(g0, [self.config.batch_size, 1])
            g0 = self.eqn.symmetrise(g0)
            v0 = y0 + self.config.delta_t \
                        * self.eqn.f(xnow, y0, z0, g0)
        else:
            g_now = self.model(x[:, :-1, :], training, 'g', idx)

        g_now = self.eqn.symmetrise(g_now)

        loss += tf.reduce_mean((g_now-tf.stop_gradient(g_target)) ** 2)
        # we are done here for fully nonlinear model
        return loss, v0


########## Section configuration ####################
class Config:
    """Configurations for defining the problem and the solver."""
    def __init__(self, dim, T, N, dtype, batch_size, train_steps,
            lr_boundaries, lr_values, eqn_name, var_reduction,
            y_neurons, z_neurons, g_neurons):
        self.dim = dim
        self.T = T
        self.N = N
        self.dtype = dtype
        self.delta_t = self.T/self.N
        self.sqrt_delta_t = np.sqrt(self.delta_t)
        self.batch_size = batch_size
        self.train_steps = train_steps
        self.lr_boundaries = lr_boundaries
        self.lr_values = lr_values
        self.eqn_name = eqn_name
        self.var_reduction = var_reduction
        self.y_neurons = y_neurons
        self.z_neurons = z_neurons
        self.g_neurons = g_neurons


########## Section main ####################
def main(eqn_name):
    T = 0.1
    N = 10
    # float64 for a better precision, float32 for smaller memory
    dtype = tf.float32

    batch_size = 256
    train_steps = 900
    lr_boundaries = [2*train_steps//3, 5*train_steps//6]
    lr_values = [0.1, 0.01, 0.001]

    var_reduction = True

    expr_name = ''
    if not var_reduction: expr_name += 'no_'
    expr_name += 'var_reduction_'
    expr_name += eqn_name
    expr_name += '.csv'

    _file = open(expr_name, 'w')
    _file.write('d,T,N,run,y0,runtime\n')

    for d in [1, 10, 100]:
        y_neurons = [d+10, d+10, 1]
        z_neurons = [d+10, d+10, d]
        g_neurons = [d+10, d+10, d*d]

        config = Config(d, T, N, dtype, batch_size, train_steps,
                    lr_boundaries, lr_values, eqn_name, var_reduction,
                    y_neurons, z_neurons, g_neurons)
        eqn = globals()[eqn_name](config)

        # 10 independent runs
        for run in range(10):
            # run on CPU to obtain reproducible results
            tf.random.set_seed(run)

            ppde_solver = PPDESolver(config, eqn)
            t_0 = time.time()
            ppde_solver.train()
            t_1 = time.time()
            _file.write('%i, %f, %i, %i, %f, %f\n'
                        % (d, T, N, run, ppde_solver.v0, t_1 - t_0))
            print(d, T, N, run, ppde_solver.v0, t_1 - t_0)
            del ppde_solver
        del config, eqn
    _file.close()

if __name__ == '__main__':
    # choice of ControlProblem, AsianOption, and BarrierOption
    # for other problems, add a new class under Section equation
    main('ControlProblem')
    main('AsianOption')
    main('BarrierOption')
\end{lstlisting}


\begin{thebibliography}{26}
\providecommand{\natexlab}[1]{#1}
\providecommand{\url}[1]{\texttt{#1}}
\expandafter\ifx\csname urlstyle\endcsname\relax
  \providecommand{\doi}[1]{doi: #1}\else
  \providecommand{\doi}{doi: \begingroup \urlstyle{rm}\Url}\fi

\bibitem[Alanko and Avellaneda(2013)]{alanko2013reducing}
S.~Alanko and M.~Avellaneda.
\newblock Reducing variance in the numerical solution of {BSDEs}.
\newblock \emph{C. R. Math. Acad. Sci. Paris}, 351\penalty0 (3-4):\penalty0
  135--138, 2013.

\bibitem[Beck et~al.(2019)Beck, Becker, Cheridito, Jentzen, and
  Neufeld]{beck2019deep}
C.~Beck, S.~Becker, P.~Cheridito, A.~Jentzen, and A.~Neufeld.
\newblock Deep splitting method for parabolic {PDEs}.
\newblock Preprint arXiv:1907.03452, 2019.

\bibitem[Dupire(2009)]{dupire2009functional}
B.~Dupire.
\newblock Functional {I}t{\^o} calculus.
\newblock Available at SSRN: https://ssrn.com/abstract=1435551 or
  https://dx.doi.org/10.2139/ssrn.1435551, 2009.

\bibitem[Ekren et~al.(2014)Ekren, Keller, Touzi, and Zhang]{ekren2014viscosity}
I.~Ekren, Ch. Keller, N.~Touzi, and J.~Zhang.
\newblock On viscosity solutions of path dependent {PDE}s.
\newblock \emph{Ann. Probab.}, 42\penalty0 (1):\penalty0 204--236, 2014.

\bibitem[Ekren et~al.(2016{\natexlab{a}})Ekren, Touzi, and
  Zhang]{ekren2016viscosity1}
I.~Ekren, N.~Touzi, and J.~Zhang.
\newblock Viscosity solutions of fully nonlinear parabolic path dependent
  {PDE}s: Part {I}.
\newblock \emph{Ann. Probab.}, 44\penalty0 (2):\penalty0 1212--1253,
  2016{\natexlab{a}}.

\bibitem[Ekren et~al.(2016{\natexlab{b}})Ekren, Touzi, and
  Zhang]{ekren2016viscosity2}
I.~Ekren, N.~Touzi, and J.~Zhang.
\newblock Viscosity solutions of fully nonlinear parabolic path dependent
  {PDE}s: Part {II}.
\newblock \emph{Ann. Probab.}, 44\penalty0 (4):\penalty0 2507--2553,
  2016{\natexlab{b}}.

\bibitem[Fahim et~al.(2011)Fahim, Touzi, and Warin]{fahim2011probabilistic}
A.~Fahim, N.~Touzi, and X.~Warin.
\newblock A probabilistic numerical method for fully nonlinear parabolic
  {PDE}s.
\newblock \emph{Ann. Appl. Probab.}, 21\penalty0 (4):\penalty0 1322--1364,
  2011.

\bibitem[Glorot and Bengio(2010)]{glorot2010understanding}
X.~Glorot and Y.~Bengio.
\newblock Understanding the difficulty of training deep feedforward neural
  networks.
\newblock In \emph{Proceedings of the thirteenth international conference on
  artificial intelligence and statistics}, pages 249--256, 2010.

\bibitem[Gobet et~al.(2005)Gobet, Lemor, and Warin]{gobet2005regression}
E.~Gobet, J.-Ph. Lemor, and X.~Warin.
\newblock A regression-based {M}onte {C}arlo method to solve backward
  stochastic differential equations.
\newblock \emph{Ann. Appl. Probab.}, 15\penalty0 (3):\penalty0 2172--2202,
  2005.

\bibitem[Han et~al.(2018)Han, Jentzen, and E]{han2018solving}
J.~Han, A.~Jentzen, and W.~E.
\newblock Solving high-dimensional partial differential equations using deep
  learning.
\newblock \emph{Proceedings of the National Academy of Sciences}, 115\penalty0
  (34):\penalty0 8505--8510, 2018.

\bibitem[Hornik(1991)]{hornik1991approximation}
K.~Hornik.
\newblock Approximation capabilities of multilayer feedforward networks.
\newblock \emph{Neural networks}, 4\penalty0 (2):\penalty0 251--257, 1991.

\bibitem[Hur{\'e} et~al.(2020)Hur{\'e}, Pham, and Warin]{hure2019some}
C.~Hur{\'e}, H.~Pham, and X.~Warin.
\newblock Deep backward schemes for high-dimensional nonlinear {PDE}s.
\newblock \emph{Mathematics of Computation}, 2020.

\bibitem[Ioffe and Szegedy(2015)]{ioffe2015batch}
S.~Ioffe and Ch. Szegedy.
\newblock Batch normalization: {Accelerating} deep network training by reducing
  internal covariate shift.
\newblock \emph{Preprint arXiv:1502.03167}, 2015.

\bibitem[Jacquier and Oumgari(2019)]{jacquier2019deep}
A.~Jacquier and M.~Oumgari.
\newblock Deep curve-dependent {PDE}s for affine rough volatility.
\newblock \emph{Preprint arXiv:1906.02551}, 2019.

\bibitem[Kingma and Ba(2014)]{kingma2014adam}
D.P. Kingma and J.~Ba.
\newblock Adam: {A} method for stochastic optimization.
\newblock \emph{Preprint arXiv:1412.6980}, 2014.

\bibitem[Peng(2011)]{peng2011note}
S.~Peng.
\newblock Note on viscosity solution of path-dependent {PDE} and g-martingales.
\newblock \emph{Preprint arXiv:1106.1144}, 2011.

\bibitem[Privault(2022)]{privaultbkf2}
N.~Privault.
\newblock \emph{Introduction to Stochastic Finance with Market Examples}.
\newblock Financial Mathematics Series. Chapman \& Hall/CRC, second edition,
  2022.

\bibitem[Ren and Tan(2017)]{ren2017convergence}
Z.~Ren and X.~Tan.
\newblock On the convergence of monotone schemes for path-dependent {PDE}s.
\newblock \emph{Stochastic Process. Appl.}, 127\penalty0 (6):\penalty0
  1738--1762, 2017.

\bibitem[Ren et~al.(2017)Ren, Touzi, and Zhang]{ren-touzi-zhang}
Z.~Ren, N.~Touzi, and J.~Zhang.
\newblock Comparison of viscosity solutions of fully nonlinear degenerate
  parabolic path-dependent {PDE}s.
\newblock \emph{SIAM J. Math. Anal.}, 49\penalty0 (5):\penalty0 4093--4116,
  2017.

\bibitem[Rogers and Shi(1995)]{rogersshi}
L.C.G. Rogers and Z.~Shi.
\newblock The value of an {A}sian option.
\newblock \emph{J. Appl. Probab.}, 32\penalty0 (4):\penalty0 1077--1088, 1995.

\bibitem[Sabate-Vidales et~al.(2020)Sabate-Vidales, {\v{S}}i{\v{s}}ka, and
  Szpruch]{sabate2020solving}
M.~Sabate-Vidales, D.~{\v{S}}i{\v{s}}ka, and L.~Szpruch.
\newblock Solving path dependent {PDE}s with {LSTM} networks and path
  signatures.
\newblock Preprint arXiv:2011.10630, 2020.

\bibitem[Saporito and Zhang(2020)]{saporito2020pdgm}
Y.F. Saporito and Z.~Zhang.
\newblock {PDGM}: {A} neural network approach to solve path-dependent partial
  differential equations.
\newblock Preprint arXiv:2003.02035, 2020.

\bibitem[Sirignano and Spiliopoulos(2018)]{sirignano2018dgm}
J.~Sirignano and K.~Spiliopoulos.
\newblock {DGM}: {A} deep learning algorithm for solving partial differential
  equations.
\newblock \emph{Journal of Computational Physics}, 375:\penalty0 1339--1364,
  2018.

\bibitem[Tang and Zhang(2015)]{tang2015}
S.~Tang and F.~Zhang.
\newblock Path-dependent optimal stochastic control and viscosity solution of
  associated {B}ellman equations.
\newblock \emph{Discrete Contin. Dyn. Syst.}, 35\penalty0 (11):\penalty0
  5521--5553, 2015.

\bibitem[Viens and Zhang(2019)]{viens2019martingale}
F.~Viens and J.~Zhang.
\newblock A martingale approach for fractional {Brownian} motions and related
  path dependent {PDE}s.
\newblock \emph{Ann. Appl. Probab.}, 29\penalty0 (6):\penalty0 3489--3540,
  2019.

\bibitem[Zhang and Zhuo(2014)]{zhang2014monotone}
J.~Zhang and J.~Zhuo.
\newblock Monotone schemes for fully nonlinear parabolic path dependent {PDE}s.
\newblock \emph{Journal of Financial Engineering}, 1\penalty0 (1):\penalty0
  1450005, 2014.

\end{thebibliography}
\end{document}